\documentclass{article}

\usepackage{microtype}
\usepackage{graphicx}
\usepackage{booktabs} 

\usepackage{hyperref}

\usepackage[algo2e]{algorithm2e} 

\usepackage[accepted]{icml2023}

\usepackage{amsmath}
\usepackage{amssymb}
\usepackage{mathtools}
\usepackage{amsthm}

\usepackage[capitalize,noabbrev]{cleveref}

\theoremstyle{plain}

\theoremstyle{definition}

\theoremstyle{remark}

\usepackage[textsize=tiny]{todonotes}

\usepackage[utf8]{inputenc} \usepackage[T1]{fontenc}    \usepackage{hyperref}       \usepackage{url}            \usepackage{booktabs}       \usepackage{amsfonts}       \usepackage{nicefrac}       \usepackage{microtype}      

\usepackage{times}
\usepackage{latexsym}
\usepackage{graphicx}
\usepackage{amsmath}
\usepackage{amsfonts}
\usepackage{graphicx}
\usepackage{color}
\usepackage{url}
\usepackage{enumitem}
\usepackage[font={footnotesize}]{caption}

\usepackage{subcaption}
\usepackage{multirow}
\usepackage{booktabs}
 \usepackage{amsthm}
\usepackage{amsmath}
\usepackage{graphicx}
\usepackage{xspace}
\usepackage{color}
\usepackage{pgfplots, pgfplotstable}
\usepackage{makecell}
\usepackage{enumitem}

\renewcommand{\vec}[1]{\boldsymbol{\mathbf{#1}}}

\def\bm{\vec{m}}

\def\bx{\vec{x}}

\def\Real{\mathbb{R}}

\def\lon{\lambda}
\def\lat{\phi}

\def\th{\vec{x}}

\newcommand{\lrunsuper}{\eta_{unsuper}}
\newcommand{\lr}{\eta_{super}}

\newcommand{\numresnet}{h}
\newcommand{\numneuron}{k}

\newcommand{\minscale}{r_{min}}

\newcommand{\imgclsloss}{l^{sup}}

\newcommand{\lossweight}{\beta}
\newcommand{\sampleset}{\mathbb{X}}

\newcommand{\Unlabelset}{\sampleset}
\newcommand{\Unbatch}{\Unlabelset_{(\batchsize)}}

\newcommand{\Labelset}{\overline{\sampleset}}

\newcommand{\embdim}{d}
\newcommand{\imgembdim}{d^{(I)}}

\newcommand{\negsamp}{\mathcal{N}}
\newcommand{\possamp}{\mathcal{P}}

\newcommand{\enc}{e}

\newcommand{\pemlp}{\mathbf{NN}}
\newcommand{\peffn}{\pemlp_{ffn}}

\newcommand{\coordspasphere}{\mathbb{S}}

\newcommand{\paramx}{\theta}
\newcommand{\parami}{\psi}

\newcommand{\image}{\mathbf{I}}

\newcommand{\classemb}{\mathbf{T}}
\newcommand{\numclass}{Q}
\newcommand{\classy}{y}

\newcommand{\act}{\sigma}

\newcommand{\dataratio}{\lambda}

\newcommand{\imgenc}{\mathbb{F}}
\newcommand{\imgcls}{g}
\newcommand{\imgprjw}{\mathbf{W}}
\newcommand{\imgencprj}{f}

\newcommand{\batchsize}{N}
\newcommand{\datasize}{M}

\newcommand{\landcovermap}{\mathbf{M}}

\newcommand{\neglocWeight}{\beta_{1}}
\newcommand{\simcseWeight}{\beta_{2}}

\newcommand{\batch}{B}
\newcommand{\geotag}{X}
\newcommand{\dropout}{D}
\newcommand{\negloc}{L}

\newcommand{\labelimg}{y}
\newcommand{\labelloc}{R}

\newcommand{\inbatchtag}{B}
\newcommand{\negloctag}{L}
\newcommand{\simcsetag}{D}

\newcommand{\neglocsize}{C}

\newcommand{\nce}{\mathrm{NCE}}
\newcommand{\lbi}{l_{\nce}}
\newcommand{\batchNLLloss}{\lbi^{\inbatchtag}}
\newcommand{\neglocNLLloss}{\lbi^{\negloctag}}
\newcommand{\simcseNLLloss}{\lbi^{\simcsetag}}

\newcommand{\neglocContWeight}{\alpha_{1}}
\newcommand{\simcseContWeight}{\alpha_{2}}

\newcommand{\mc}{\mathrm{MC}}
\newcommand{\lmc}{l_{\mc}}
\newcommand{\batchContloss}{\lmc^{\inbatchtag}}
\newcommand{\neglocContloss}{\lmc^{\negloctag}}
\newcommand{\simcseContloss}{\lmc^{\simcsetag}}

\newcommand{\batchTmp}{\tau_0}
\newcommand{\neglocTmp}{\tau_1}
\newcommand{\simcseTmp}{\tau_2}

\newcommand{\mse}{MSE}

\newcommand{\tile}{tile}
\newcommand{\aodha}{wrap}

\newcommand{\gridcell}{grid}
\newcommand{\theory}{theory}

\newcommand{\modelfullname}{Contrastive Spatial Pre-Training}
\newcommand{\modelname}{{CSP}}

\newcommand{\vspacepred}{\vspace*{-0.5cm}}

\newcommand{\imageonly}{Img. Only}
\newcommand{\superonly}{Sup. Only}

\newcommand{\locdim}{d}

\usepackage{amsmath,amsfonts,bm}

\def\eqref#1{equation~\ref{#1}}

\def\1{\bm{1}}

\def\rva{{\mathbf{a}}}
\def\rvb{{\mathbf{b}}}

\DeclareMathAlphabet{\mathsfit}{\encodingdefault}{\sfdefault}{m}{sl}
\SetMathAlphabet{\mathsfit}{bold}{\encodingdefault}{\sfdefault}{bx}{n}

\newcommand{\sigmoid}{\sigma}

\icmltitlerunning{\modelname: Self-Supervised \modelfullname{} for Geospatial-Visual Representations}

\begin{document}

\twocolumn[
\icmltitle{\modelname: Self-Supervised \modelfullname{}\\ for Geospatial-Visual Representations}

\icmlsetsymbol{equal}{*}

\begin{icmlauthorlist}
\icmlauthor{Gengchen Mai}{equal,uga,stanford,ugacs}
\icmlauthor{Ni Lao}{equal,google}
\icmlauthor{Yutong He}{stanford,cmu}
\icmlauthor{Jiaming Song}{stanford}
\icmlauthor{Stefano Ermon}{stanford}
\end{icmlauthorlist}

\icmlaffiliation{uga}{Spatially Explicit Artificial Intelligence Lab, Department of Geography, University of Georgia, USA}
\icmlaffiliation{ugacs}{School of Computing, University of Georgia, USA}
\icmlaffiliation{google}{Google Inc, USA}
\icmlaffiliation{stanford}{Department of Computer Science, Stanford University, USA}
\icmlaffiliation{cmu}{Machine Learning Department, Carnegie Mellon University, USA}

\icmlcorrespondingauthor{Gengchen Mai}{gengchen.mai25@uga.edu}
\icmlcorrespondingauthor{Ni Lao}{nlao@google.com}

\icmlkeywords{Constrastive Learning, Location Encoder, Self-Supervised Learning}

\vskip 0.3in
]

\printAffiliationsAndNotice{\icmlEqualContribution} 

\begin{abstract}

Geo-tagged images are publicly available in large quantities, whereas labels such as object classes 
are rather scarce and expensive to collect. 
Meanwhile, contrastive learning has achieved tremendous success in various 
natural image and language 
tasks with limited labeled data.
However, existing methods  fail to fully leverage geospatial information, which can be paramount to distinguishing objects that are visually similar.
To directly leverage the abundant geospatial information associated with images 
in pre-training, fine-tuning, and 
inference stages,
we present \modelfullname~(\modelname), a self-supervised learning framework for geo-tagged images.
We use a dual-encoder to separately encode the images and their corresponding geo-locations, and use 
contrastive objectives to learn effective location representations from images, which can be transferred  to downstream supervised tasks such as image classification. 
Experiments 
show that \modelname~can improve model performance on both iNat2018 and fMoW dataset. Especially, on iNat2018, \modelname{}
significantly boosts the model performance with 10-34\% relative improvement with various labeled training data sampling ratios\footnote{Code, data, and pre-trained models are available at \url{https://gengchenmai.github.io/csp-website/}.}.

\end{abstract} 
\vspace{-0.1cm}
\section{Introduction}
\vspace{-0.1cm}
\label{sec:intro}

\begin{figure}[t!]
	\centering \begin{subfigure}[b]{0.13\textwidth}  
		\centering 
		\includegraphics[width=\textwidth]{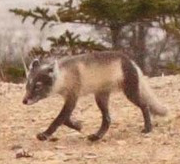}\vspace*{-0.1cm}
		\caption[]{{\footnotesize 
		Arctic Fox
		}}    
		\label{fig:fox1}
	\end{subfigure}
	\hfill
	\begin{subfigure}[b]{0.32\textwidth}  
		\centering 
		\includegraphics[width=\textwidth]{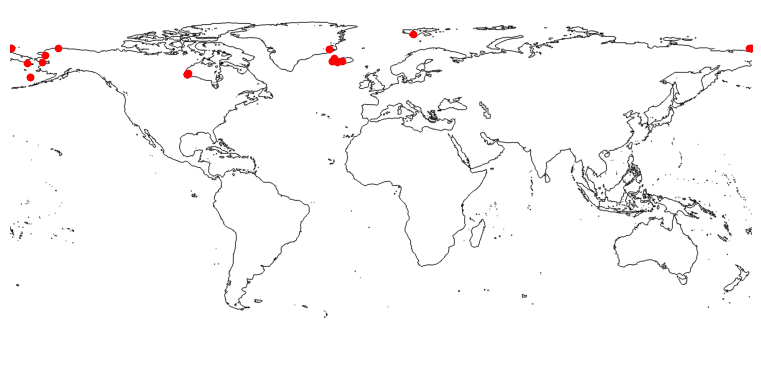}\vspacepred
	\vspace*{-0.1cm}
		\caption[]{{\footnotesize 
		Arctic Fox Locations }}    
		\label{fig:4084_dist_intro}
	\end{subfigure}
	
	\begin{subfigure}[b]{0.13\textwidth}  
		\centering 
		\includegraphics[width=\textwidth]{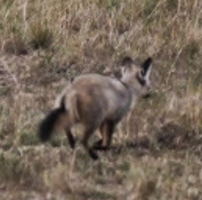}\vspace*{-0.1cm}
		\caption[]{{\footnotesize 
		Bat-Eared Fox
		}}    
		\label{fig:fox2}
	\end{subfigure}
	\hfill
	\begin{subfigure}[b]{0.32\textwidth}  
		\centering 
		\includegraphics[width=\textwidth]{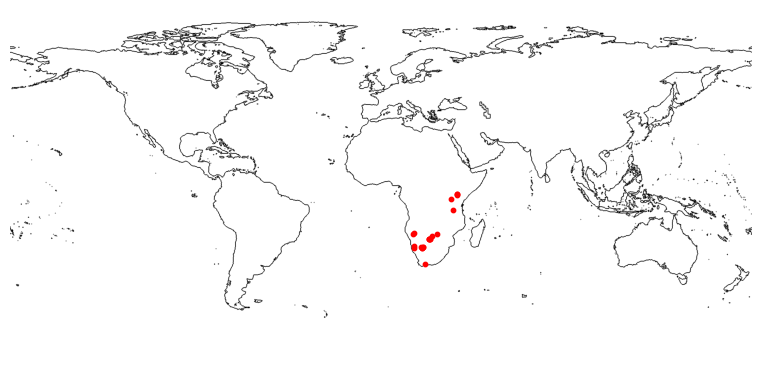}\vspacepred
	\vspace*{-0.1cm}
		\caption[]{{\footnotesize 
		Bat-Eared Fox Locations}}    
		\label{fig:4081_dist_intro}
	\end{subfigure}
	\vspace{-0.2cm}
	\caption{
	The importance of geospatial information
demonstrated by
	two visually similar species (a)(c), and
their distinct patterns in  image locations (b)(d). 
} 
	\label{fig:spesdist_intro}
 	\vspace{-0.4cm}
\end{figure}

\begin{figure*}[t!]
	\centering \begin{subfigure}[b]{0.35\textwidth}  
		\centering 
		\includegraphics[width=\textwidth]{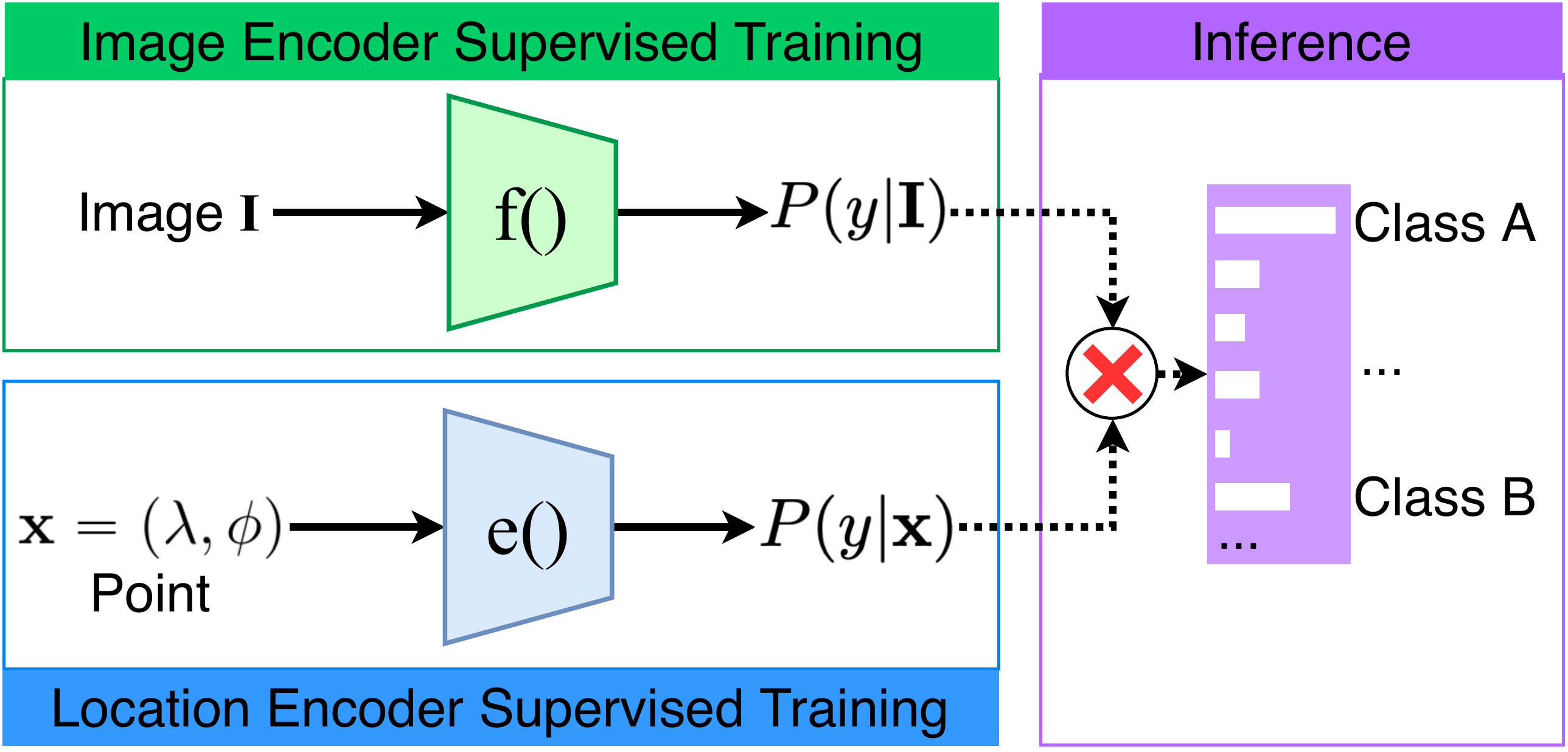}
		\caption[]{{\textbf{\superonly}: Geo-aware Supervised Learning \cite{mac2019presence,mai2020multiscale}
		}}    
		\label{fig:model-geo-super}
	\end{subfigure}
	\hfill \hfill
	\begin{subfigure}[b]{0.62\textwidth}  
		\centering 
		\includegraphics[width=\textwidth]{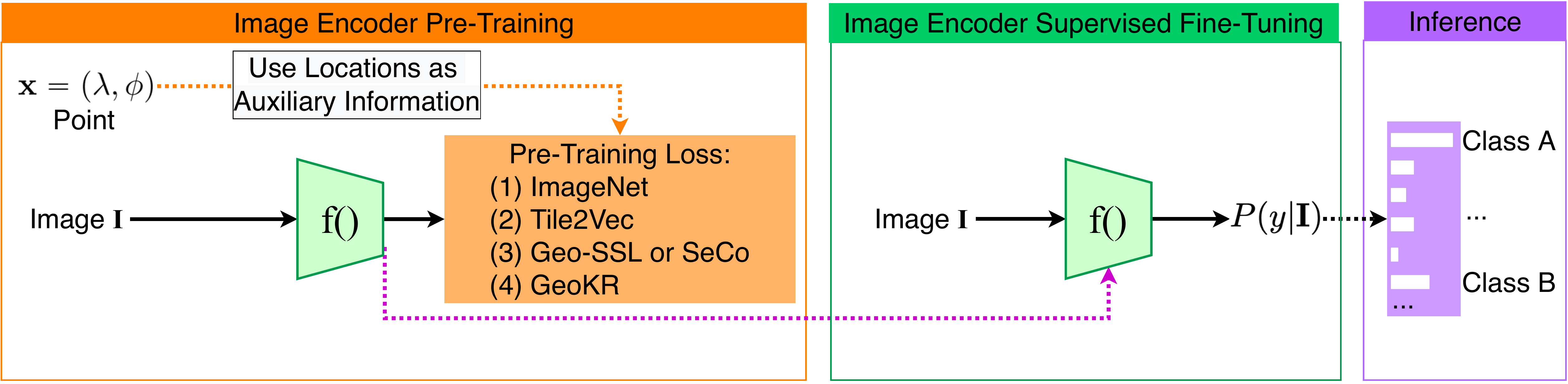}
		\caption[]{{\textbf{\imageonly}: Image Encoder Pre-Training with Geographic Knowledge \cite{jean2019tile2vec,ayush2020selfsup,manas2021seasonal,li2021geographical}.
  Here we show four previous approaches to pre-train the image encoder $\imgencprj()$ (orange box). A detailed version can be seen in Figure \ref{fig:model-geo-ssl-detail} in Appendix \ref{sec:mode-geo-ssl-fig}.
		}}    
		\label{fig:model-geo-ssl}
	\end{subfigure}
	\hfill
	\vspace{0.2cm}
	\begin{subfigure}[b]{1.0\textwidth}  
		\centering 
		\includegraphics[width=\textwidth]{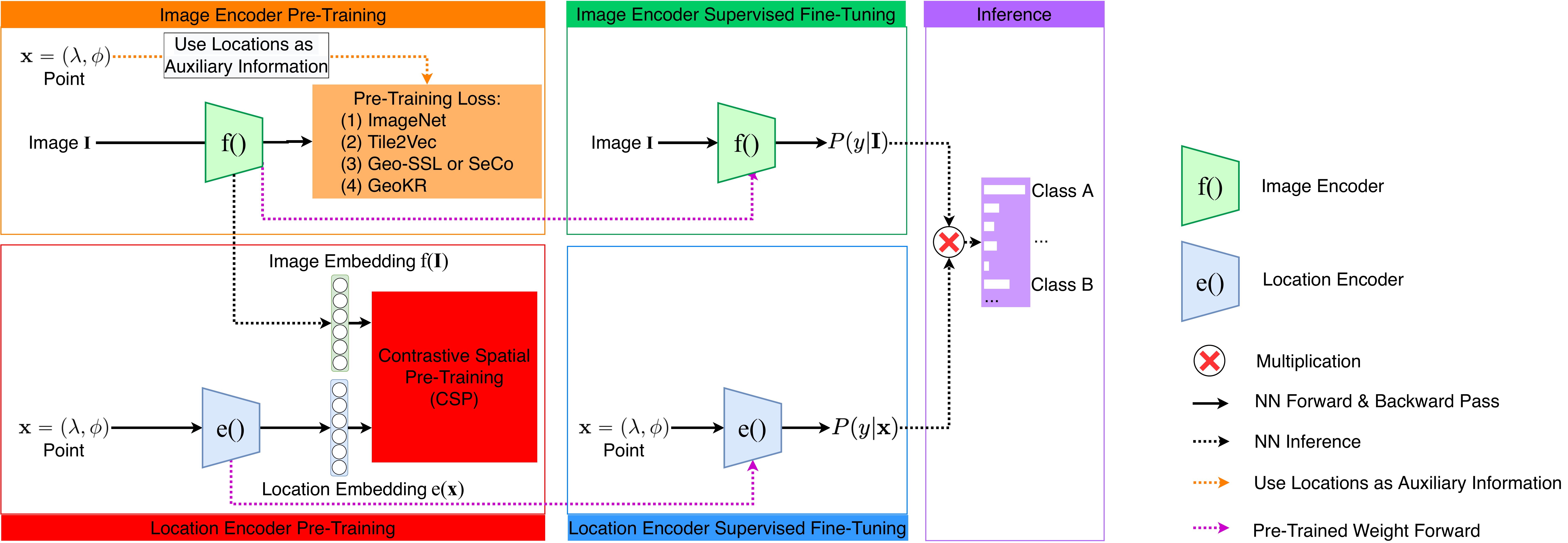}
		\caption[]{{\textbf{\modelfullname~ (\modelname)}.
  It adds a location encoder pre-training stage (red box). 
	The pre-trained $\imgencprj()$ is used as an unsupervised feature extractor to generate image embeddings $\imgencprj(\image)$ which are used for contrastive learning with location embedding $\enc(\th)$. 
	Both $\imgencprj()$ and $\enc()$ are fine-tuned in a supervised manner (green and blue box).
		}}    
		\label{fig:model-csp}
	\end{subfigure}
	\caption{
Different training strategies for geo-aware image classification. Our proposed method \modelname~is presented in Figure \ref{fig:model-csp}.
} 
	\label{fig:model_illus}
 	\vspace{-0.4cm}
\end{figure*}

Low-data or few-shot regimes \cite{zhai2021lit, wang2020fsl} is a prevalent challenge in the geospatial domain, where we usually have access to massive amounts of unlabeled data while only limited amount of labeled data is available.
For example, users on Flickr, Google Photos, and iNaturalist App\footnote{iNaturalist is one of the world's most popular nature apps to help users identify species given the uploaded images.} upload millions of geo-tagged images every day, and multiple satellites continuously capture remote sensing (RS) images with corresponding geo-coordinates all over the world. These geo-tagged data form large publicly available \textit{unlabeled} datasets that are inexpensive to obtain. 
In contrast, desired labels for many geospatial tasks (e.g., object class labels, object bounding boxes, and land use type labels, etc.) are rather scarce and expensive to collect.
Moreover, even well-curated and widely used labeled geospatial datasets such as UC Merced Land Use Dataset \cite{yang2010ucmerced} and BigEarthNet \cite{sumbul2019bigearthnet} have limited sizes, geographic coverage, and potentially oversimplified label distributions.
This lack of labeled data coverage severely limits the ability to generalize, especially in a geographic sense, of models trained on these labeled geospatial datasets \cite{goodchild2021replication}.

Meanwhile, numerous previous studies have shown
the great potential of leveraging geospatial information as complementary information for visual cues to help improve the model performance
on various computer vision tasks
\cite{tang2015improving,chu2019geo,mac2019presence,klocek2019functional,mai2020multiscale,mai2022sphere2vec,yang2022dynamic}.
For example, Figure \ref{fig:fox1} and \ref{fig:fox2} are images of two different fox species: Arctic fox and bat-eared fox, with which the vision-based models, or even humans, can be confused due to the high visual similarity of the two species and their surrounding environments.
Fortunately, 
these two species have distinct geospatial distribution patterns (shown in Figure~\ref{fig:4084_dist_intro}, \ref{fig:4081_dist_intro}), 
and it is very easy to tell them apart based on the geo-locations.
Motivated by these observations, we ask whether we can \textbf{build a  multi-modal self-supervised learning framework between geo-locations and images} that learns the alignments between geo-location and image representations using large unlabeled geo-tagged datasets.

In this work, we propose \modelname~(\modelfullname), 
a self-supervised learning framework, which pre-trains deep spatial representations from unlabeled geo-tagged images by predicting image features or image identities based on their geo-locations as shown in Figure \ref{fig:model-csp}.
Given one location-image pair
$(\th_i, \image_i)$, a dual-encoder separately encodes $\th_i$ and $\image_i$ into the embedding space with a location encoder $\enc()$ and an image encoder $\imgencprj()$ and contrast against related locations and images to form a \textbf{contrastive learning objective} (the red box in Figure \ref{fig:model-csp}). After the location encoder and image encoder pre-training stage, both $\enc()$ and $\imgencprj()$ can be fine-tuned on a small amount of labeled data (the green and blue box in Figure \ref{fig:model-csp}) separately and do inference jointly, which is compatible with prior works \cite{mac2019presence,mai2020multiscale}.

To perform contrastive learning, we explore a combination of three different ways to form positive and negative pairs for the location encoder pre-training stage of \modelname~as shown in Figure \ref{fig:csp_loss}: 
\textbf{(a) In-batch negative sampling}: given a mini-batch of unlabeled location-image pairs, create mismatching location-image pairs as negative samples;
\textbf{(b) Random negative location sampling}:  uniformly sample negative locations from the study area (e.g., the whole earth surface) to form negative pairs;
\textbf{(c) SimCSE-based sampling}:  create a positive pair by encoding the same location with two location encoders,  which share all the parameters but use different dropout masks.
We also compare several self-supervised learning objectives including 
\textbf{Mean Square Error} loss  ($\mse$), 
\textbf{Noise Contrastive Estimation} loss ($\nce$), and
\textbf{Contrastive Multi-classification} loss ($\mc$).

We conduct experiments on geo-aware image classification tasks including \textbf{fine-grained species recognition} \cite{chu2019geo,mac2019presence,mai2020multiscale,yang2022dynamic}, 
and \textbf{remote sensing (RS) image classification}  \cite{christie2018functional,ayush2020selfsup,manas2021seasonal,li2021geographical}. 
Results show that our \modelname~can boost the model performance on both datasets.

\textbf{In summary, the contributions of our work are:}
\begin{itemize}[leftmargin=*]\item We propose an effective multi-modal self-supervised pre-training method \modelname~that leverages abundant unlabeled geo-tagged images to better learn location representations that can be transferred to few-shot learning tasks. 
    \item We explore three ways to construct 
positive and negative training examples for contrastive learning. We find that the combination of them achieves the best performance. 
\item We explore three self-supervised losses including $\mse$, $\nce$, and $\mc$. We find out that using \modelname{} with $\mc$ usually yields the best result.
    \item  We apply \modelname~to fine-grained species recognition  (iNat2018) 
and remote sensing image classification task (fMoW) in few-shot learning and fully supervised settings, and demonstrate advantages on both datasets. \modelname{} can significantly boost model performances with 10-34\% relative improvements on the iNat2018 dataset at few-shot settings by stratified sampling $\{5\%, 10\%, 20\%\}$ of the training data. On both datasets, when training models on the whole training dataset in a fully supervised manner, we find that adding the \modelname~ pre-training objective can still improve the model performance. \end{itemize}

 \section{Related Work}
\label{sec:relatedwork}

\paragraph{Unsupervised/Self-Supervised Learning on Geotagged Images}
Multiple unsupervised or self-supervised frameworks have been proposed to pre-train image encoder by utilizing geographic knowledge such as Tile2Vec \cite{jean2019tile2vec}, Geo-SSL \cite{ayush2020selfsup}, SeCo \cite{manas2021seasonal}, and GeoKR \cite{li2021geographical}. 

\textbf{Tile2Vec} \cite{jean2019tile2vec} is an unsupervised learning framework to pre-train image encoder based on the spatial relations among RS images. Given an anchor RS image, location information is only used to obtain one nearby tile and a distant tile. An unsupervised triplet loss is formed to pre-train image encoder to make nearby tiles similar in the embedding space while distant tiles dissimilar. 
Geo-locations are not part of the model input and cannot be used during the model fine-tuning or inference stage.

\textbf{Geo-SSL} \cite{ayush2020selfsup} is a self-supervised contrastive learning objective to pre-train an RS image encoder based on the MoCo-V2 \cite{chen2020improved} framework. Instead of using augmented images as positive pairs as MoCo-V2 does, they used co-located RS images at different times as positive pairs. 
This contrastive image loss is combined with a geo-location classification pre-text loss during pre-training, which
uses the image encoder 
to predict which geo-location cluster the image might come from. 
Here, the spatiotemporal information is only used in the pre-training stage. During the fine-tuning and inference stage, the model prediction relies entirely on the pre-trained image encoder.

\textbf{SeCo} \cite{manas2021seasonal} is a similar self-supervised contrastive learning framework for an RS image encoder $\imgencprj()$. It also uses MoCo-V2 as the backbone and uses spatially aligned RS images at different times as novel temporal augmented samples. The difference is that SeCo  
uses both the temporal augmented samples and synthetic samples based on artificial augmentations as either positive or negative samples so that the pre-trained $\imgencprj()$ can be either invariant or sensitive to the temporal or artificial augmentations.

\textbf{GeoKR} \cite{li2021geographical} is proposed as an unsupervised framework for an RS image encoder. GeoKR first obtains a spatially aligned land cover map $\landcovermap$ 
based on an RS image. The image encoder is pre-trained in a teacher-student network to predict the distribution of land cover types in the current scene with a KL loss. 

Figure \ref{fig:model-geo-ssl} illustrates the general idea of those four models while
Figure \ref{fig:model-geo-ssl-detail} in Appendix \ref{sec:mode-geo-ssl-fig} provides a detailed comparison. 
None of them directly takes geo-locations as model input but use locations as auxiliary information to pre-train the image encoder. Moreover, after pre-training, location information is completely ignored during fine-tuning and inference stage which leads to significantly suboptimal results.
In contrast, our \modelname~utilizes the location-image pairs in a direct and explicit manner by separately encoding them and contrasting them against each other. The pre-trained location encoder can be utilized in the model inference process jointly with the image encoder so that both the visual and spatial clue can be used for prediction.

\paragraph{Location Representation Learning}
\citet{zhai2018geotemporal} learned location representation from image-location pairs for image localization. So in this context, locations are supervision signals. 
Instead of using the original geo-locations, they grouped locations (or times) into different bins and utilized them in the cross entropy loss. This practice cannot leverage the continuity of the approximated function.
Most existing location encoding approaches
\cite{tang2015improving,christie2018functional,chu2019geo,mac2019presence,mai2020multiscale,mai2022sphere2vec,yang2022dynamic} are developed and trained in a supervised learning framework while massive unlabeled geographic data cannot be used. 
Figure \ref{fig:model-geo-super} illustrates the dual-encoder supervised learning idea both \citet{mac2019presence} and \citet{mai2020multiscale} used for geo-aware image classification. 
In contrast, this work focuses on training location encoders in a self-supervised manner based on unlabeled geotagged images. The pre-trained location encoder can later be utilized jointly with the image encoder for model prediction (See Figure \ref{fig:model-csp}). 

\paragraph{Spatially Explicit Artificial Intelligence}
Previous works showed that naively applying existing state-of-the-art AI models to geospatial tasks  usually yielded suboptimal results \citep{mai2019relaxing,chu2019geo,mac2019presence,ayush2020selfsup,yan2018xnet+}. 
\textit{Spatially Explicit Artificial Intelligence} aims at improving the performances of AI models on various geospatial tasks by incorporating spatial thinking, spatial principles and spatial inductive biases into the AI model design \citep{janowicz2020geoai,liu2022review,zhu2022reasoning,mai2022symbolic}. 
Several important spatial principles have been considered by previous works including spatial dependency \citep{mai2019relaxing,yan2018xnet+,yan2019spatial,li2021tobler,huang2022estimating,huang2023learning}, spatial heterogeneity \citep{chu2019geo,mac2019presence,mai2020multiscale,zhu2021spatial,xie2021statistically,goodchild2021replication,xie2023harnessing}, temporal continuity \citep{cai2020traffic,he2021spatial,cong2022satmae}, temporal periodicity \citep{cai2020traffic,rao2020lstm}, 
earth spherical geometry nature \citep{cohen2018spherical,esteves2018learning,jiang2019spherical,mai2022sphere2vec}, and so on.
\modelname~  contributes to the \textit{Spatially Explicit Artificial Intelligence} research by learning effective multi-scale location representations from unlabeled images.

 \section{Method} \label{sec:method}

\begin{figure*}[t!]
	\centering 

	\vspace*{-0.2cm}
	\begin{subfigure}[b]{0.3\textwidth}  
		\centering 
		\includegraphics[width=\textwidth]{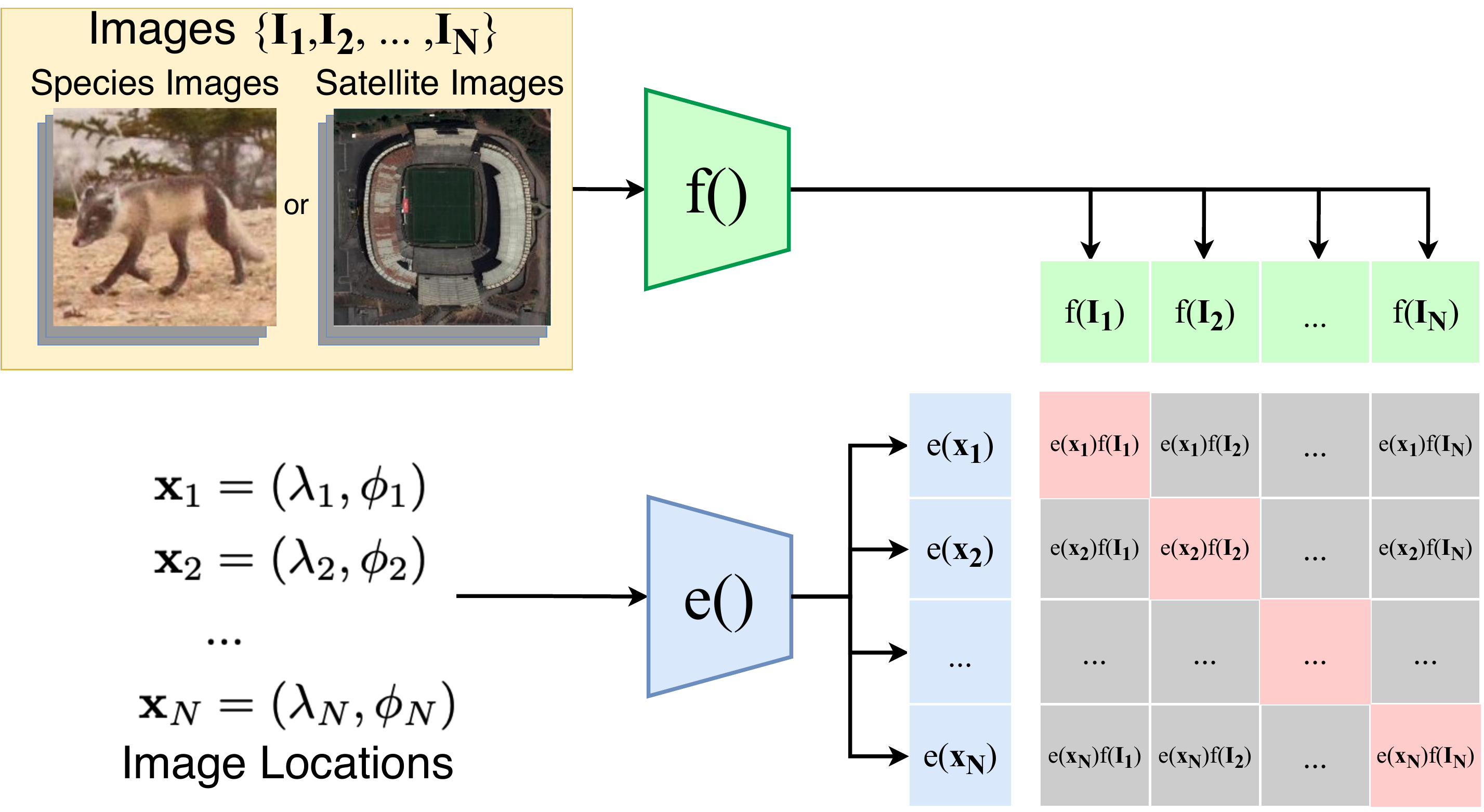}\vspace{0.5cm}
		\caption[]{{ 
		\textbf{In-batch negative sampling ($\inbatchtag$)}
  Given a batch of unlabeled location-image pairs a $\batchsize \times \batchsize$ cosine similarity matrix is computed based on all location embeddings and image embeddings. 
		}}    
		\label{fig:inbatch_loss}
	\end{subfigure}
\hspace{0.03\textwidth}
	\begin{subfigure}[b]{0.3\textwidth}  
		\centering 
		\includegraphics[width=\textwidth]{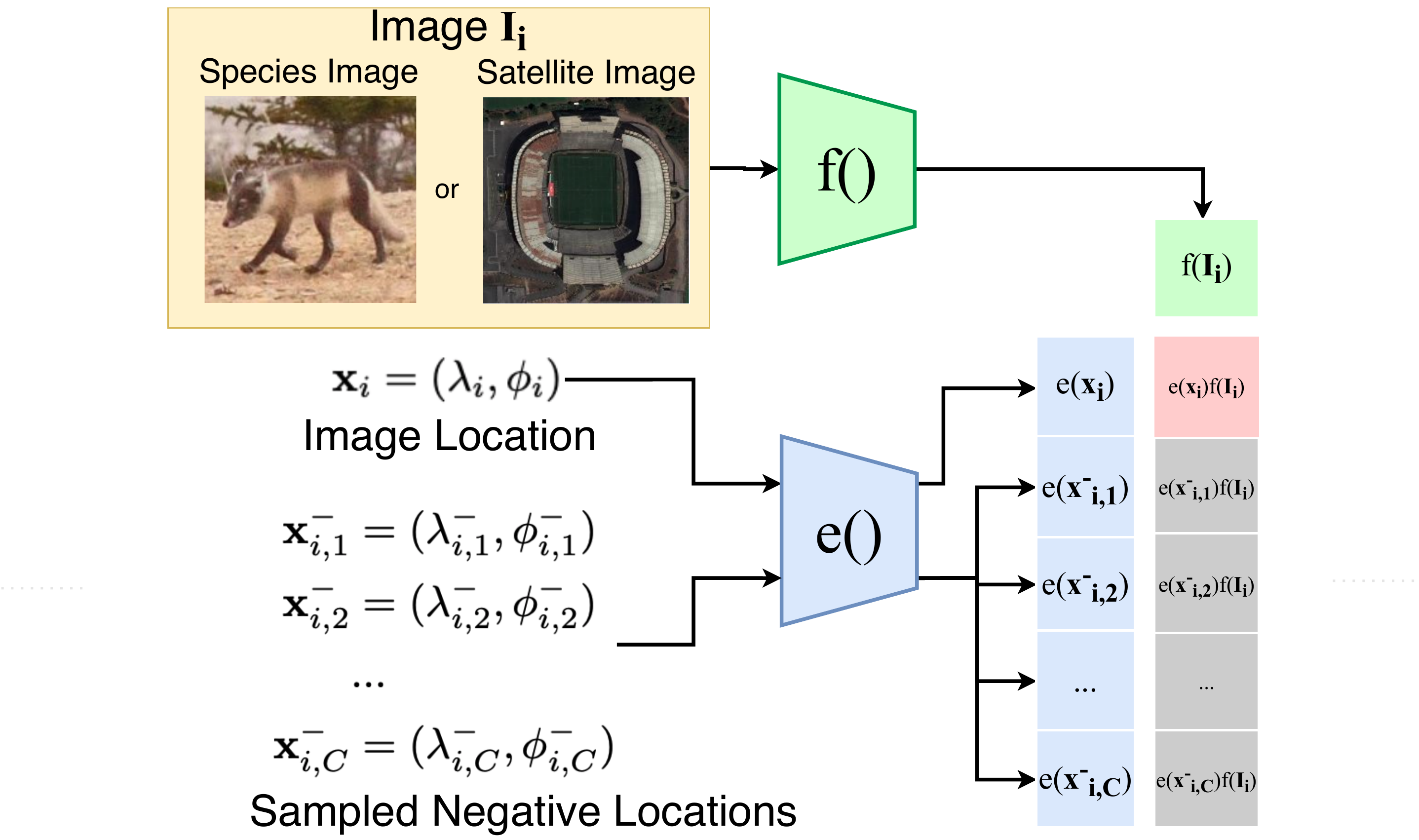}\vspace{-0.1cm}
		\caption[]{{ \textbf{Random negative location sampling ($\negloctag$)}
Given one positive location-image pair $(\th_i, \image_i)$, we uniformly sample $\neglocsize$ negative locations $\{\th^{-}_{i,1}, \th^{-}_{i,2},...,\th^{-}_{i,\neglocsize}\}$ from the study area (e.g., the earth surface) to form negative pairs.
		}}    
		\label{fig:negloc_loss}
	\end{subfigure}
\hspace{0.03\textwidth}
	\begin{subfigure}[b]{0.31\textwidth}  
		\centering 
		\includegraphics[width=\textwidth]{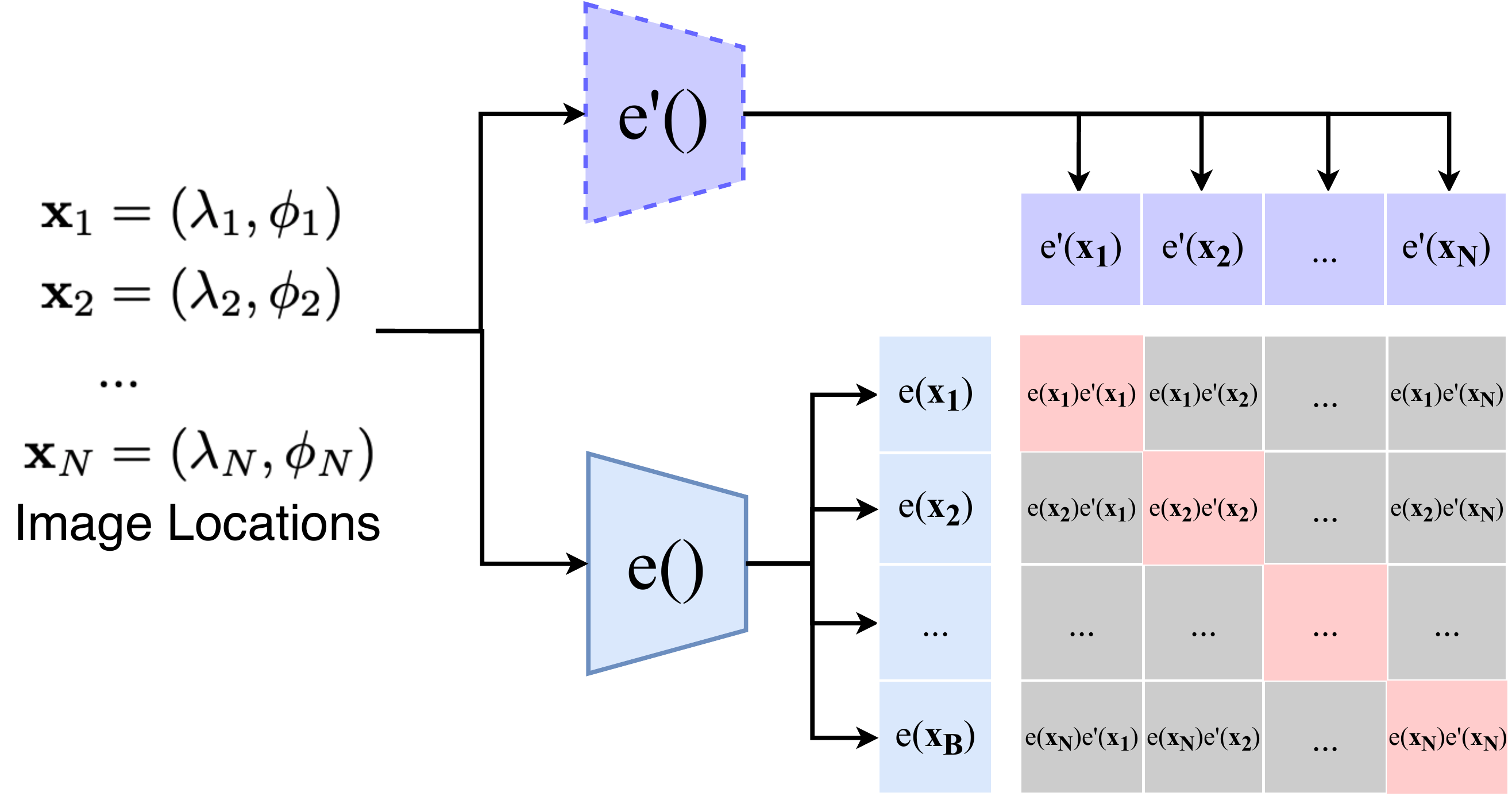}
		\vspace{-0.4cm}
		\caption[]{{ 
		\textbf{SimCSE sampling ($\simcsetag$)}
  Two location encoders $\enc()$ and $\enc'()$ share all the parameters but apply different dropout masks.
Given a batch of unlabeled location-image pairs, the same location $\th_i$ is encoded by $\enc()$ and $\enc'()$ to form a positive pair, while other pairs are treated as negative examples.
		}}    
		\label{fig:simcse_loss}
	\end{subfigure}
	\vspace*{-0.1cm}
	\caption{
Three different ways to form positive and negative training pairs 
(red and gray boxes respectively). 
} 
	\label{fig:csp_loss}
\end{figure*}

\subsection{A Dual-Encoder for Geo-Tagged Images}  \label{sec:dual_enc}

We define an unlabeled geo-tagged image dataset as
$\Unlabelset = \{(\th_i, \image_i)|i = 1, ...,\datasize\}$, where $\image_i$ is an image, $\th_i$
represents the location (longitude and latitude) and optionally the time 
the image was taken\footnote{In this study we focus on the location information and leave the time aspect to the future work.}. 
Inspired by recent image-text pre-training models \cite{zhang2020contrastive,radford2021clip,jia2021scaling,zhai2021lit}, 
\modelname~uses a dual-encoder architecture -- a location encoder $\enc()$ and an image encoder $\imgencprj()$ -- to handle location $\th_i$ and image $\image_i$ separately.

The location encoder $\enc()$ is  a function 
$\enc_{\paramx}(\th_i): \coordspasphere^{2} \to \Real^\embdim$,
which is parameterized by $\paramx$ and maps any coordinate $\th_i=(\lon_i,\lat_i)$ in a spherical surface $\coordspasphere^{2}$ to a vector representation of $\embdim$ dimension.
Here longitude $\lon_i \in  [-\pi , \pi)$ and latitude $\lat_i \in [-{\pi}/{2} , {\pi}/{2}]$.
$\enc()$ can be any existing 2D location encoders \cite{mai2022review} such as $\tile$ \cite{tang2015improving}, $\aodha$ \cite{mac2019presence}, Space2Vec's $\gridcell$ and $\theory$ \cite{mai2020multiscale}, or spherical location encoders such as Sphere2Vec \cite{mai2022sphere2vec}.
We assume that $\enc()$ is inductive and does not depend on the unlabeled dataset $\Unlabelset$ anymore once it is pre-trained. 

The image encoder $\imgencprj()$ is a function 
$\imgencprj_{\parami}(\image_i): \Real^{H \times W \times C} \to \Real^\embdim$, 
which is parameterized by $\parami$ and maps any image with height $H$, width $W$, and channel $C$ into an embedding of $\embdim$ dimension. 
In this study we define $\imgencprj(\image_i)=\imgprjw(\imgenc(\image_i))$ where $\imgenc()$ is an off-the-shelf deep image neural network such as InceptionV3 \cite{szegedy2016rethinking} or Geo-SSL \cite{ayush2020selfsup} pretrained ResNet50 \cite{he2015resnet}, which encodes any image into a $\imgembdim$ dimension image feature vector. 
$\imgprjw()$ is a projection layer (similar to that of SimCLR \cite{chen20simclr} and MoCo-V2 \cite{chen2020improved}), which projects the image feature $\imgenc(\image_i) \in \Real^{\imgembdim}$ into $\embdim$ dimension such that a contrastive learning objective can be formed between $\enc(\th_i)$ and $\imgencprj(\image_i)$. Please refer to Appendix \ref{sec:img_enc_img_enc_describe} for a detailed description of $\imgencprj()$.

In our work, $\imgembdim = 2048$ and $\embdim = 512$. This dual-encoder architecture is shown in Figure \ref{fig:model-csp} as well as Figure \ref{fig:csp_loss}.
We simply denote the encoded representation of a location $\th_i$  as $\enc(\th_i)$ and its associated image representation as $\imgencprj(\image_i)$.

\subsection{\modelfullname (\modelname)}
\label{sec:csp_loss}

\paragraph{Contrastive Learning Objectives} 
We consider different contrastive objectives. 
The first is the \textit{noise contrastive estimation} (NCE) \cite{gutmann2010nce} loss, which avoids calculation of the partition function and has been successfully used in word embeddings \cite{mikolov2013distributed} and language modeling \cite{mnih2012lm}:
\begin{align}
\begin{split}
    \lbi(\possamp, \negsamp) = 
    &-  \mathbb{E}_{(\rva, \rvb) \sim \possamp}\log \sigma(s(\rva, \rvb))  \\
    &
-\mathbb{E}_{(\rva, \rvb^{-}) \sim \negsamp}\log (1 - \sigma(s(\rva, \rvb^{-})))
\end{split}
\label{equ:nce}
\end{align}
Here $\possamp=\{(\rva, \rvb)\}$ is a set of positive pairs, and $\negsamp=\{(\rva, \rvb^{-})\}$ is a set of negative pairs. $s(\cdot, \cdot)$ is a similarity function 
(such as $cosine()$), and $\sigma(v) = e^v / (1 + e^v)$ is the sigmoid function. 

The second objective function is the multi-class classification loss with temperature which takes the same form as the InfoNCE loss \cite{van2018representation}. It has been successfully used in 
unsupervised learning for images 
\cite{he2020moco} and text 
\cite{gao2021simcse}:
{\begin{align}
\begin{split}\small
    \lmc(\possamp, \negsamp, \tau) \\
    = 
    \mathbb{E}_{(\rva, \rvb) \sim \possamp}&\frac{
        e^{s(\rva, \rvb)/\tau}
}{
    e^{s(\rva, \rvb)/\tau} + 
\sum_{(\rva, \rvb^{-}) \in \negsamp_\rva}
    e^{s(\rva, \rvb^{-})/\tau}
}
\end{split}
    \label{equ:mc}
\end{align}
}
where $\mc$ stands for ``multi-class''.
$\negsamp_\rva$ obtains a set of negative pairs with first entry being $\rva$, 
$\possamp$ and $s(\cdot, \cdot)$ are defined as earlier. 
The temperature scaling parameter $\tau$ determines how soft the softmax is \cite{hinton2015distilling}. In practice it helps with the trade off between top ranked classes (precision) versus reset of the classes (recall).

Third, we also experimented with a regression loss, but it does not work as well as the NCE and MC losses.

\paragraph{Self-Supervised Training Pair Construction}
In order to learn useful representations, we need to choose appropriate distributions for positive pairs $\possamp$ and negative pairs $\negsamp$ for contrastive learning. 
In \modelname{}, we use three sampling methods to obtain positive and negative pairs: in-batch negative sampling (indicated as $\inbatchtag$), random negative location sampling (indicated as $\negloctag$), and SimCSE-based sampling (indicated as $\simcsetag$).
Figure \ref{fig:csp_loss} illustrates how we use these three methods to do the positive and negative sampling. 
Each of them includes methods to sample both the positive and negative pairs so that one contrastive loss component can be formed based on each of them. 
Some of them share the same positive sampling method such as $\inbatchtag$ and $\negloctag$. So we summarize the positive and negative sampling methods below.
Given an unlabeled location-image pair $(\th_i, \image_i)$ from a mini-batch $\Unbatch = \{(\th_1, \image_1), (\th_2, \image_2),...,(\th_{\batchsize}, \image_{\batchsize})\} \subseteq \Unlabelset$, where $\Unlabelset$ is a geo-tagged but unlabeled image set,
we use the following 
positive and negative instances:
\begin{itemize}[leftmargin=*]
\item \textbf{Geo-tagged positive} $\possamp^{\geotag} = \{(\enc(\th_i), \imgencprj(\image_i) )\}$ indicates the original location-image pairs used as positive pairs. This corresponds to the positive pairs used by $\inbatchtag$ and $\negloctag$ methods -- the red boxes in both Figure \ref{fig:inbatch_loss} and \ref{fig:negloc_loss}. 
\item \textbf{In-batch negatives } $\negsamp^{\batch} = \bigcup_{i } \negsamp^{\batch}_i $, where $\negsamp^{\batch}_i = \{ (\enc(\th_i), $ $\imgencprj(\image_j)) | j \in \{1,2,...,\batchsize\}\setminus\{i\}\}$.$\negsamp^{\batch}$ corresponds to all mismatching location-image pairs in $\Unbatch$ -- all gray boxes (no-diagonal elements) in Figure \ref{fig:inbatch_loss}.
\item \textbf{Sampled negative locations}  
    $\negsamp^{\negloc} = \bigcup_{i } \negsamp^{\negloc}_i $, where
    $\negsamp^{\negloc}_i = \{( \enc(\th^{-}_{i,j}), $ $\imgencprj(\image_i) ) | j \in \{1,2,...,\neglocsize\} \}$ indicates $\neglocsize$ negative pairs for $\image_i$. 
    Note that $\th^{-}_{i,j}$ is sampled uniformly 
from the surface of the sphere at pre-training time, and therefore they are different at each training epoch.
$\negsamp^{\negloc}$ corresponds to all gray boxes in Figure \ref{fig:negloc_loss}. 
    This is a common negative location sampling practice used by \citet{mac2019presence,mai2020multiscale}.
\item \textbf{Dropout positive} $\possamp^{\dropout} = \{( \enc(\th_i), \enc'(\th_i) )\}$, where 
    given two towers of the same location encoders $\enc()$ and $\enc'()$ with two independently sampled dropout masks, 
we pass the same input $\th_i$ to them and obtain two embeddings $( \enc(\th_i), \enc'(\th_i) )$ as “positive pairs”. This is a data augmentation strategy (so called SimCSE), which has been very successful for sentence embeddings \cite{gao2021simcse}. This corresponds to the red boxes 
in Figure \ref{fig:simcse_loss}.
\item \textbf{Dropout negative} $\negsamp^{\dropout} = \bigcup_i \negsamp^{\dropout}_i$, where $\negsamp^{\dropout}_i = \{( \enc(\th_i), $ $ \enc'(\th_j) )  | j \in \{1,2,...,\batchsize\} \setminus\{j\}\}$.
$\negsamp^{\dropout}$ indicates the location embeddings from two location encoder towers based on different locations from the same mini-batch. It corresponds to the gray boxes in Figure \ref{fig:simcse_loss}. 
\end{itemize}

As shown in Figure \ref{fig:csp_loss}, those five positive/negative sampling sets amount to three different sampling methods: 
\begin{itemize}[leftmargin=*]
\item \textbf{In-batch negative sampling ($\inbatchtag$)}  \cite{zhang2020contrastive,radford2021learning,carlsson2020semantic,karpukhin2020dpr} uses $\possamp^{\geotag}, \negsamp^{\batch}$ as positive and negative pairs.
    \item \textbf{Random negative location sampling ($\negloctag$)}  \cite{mac2019presence,mai2020multiscale,mai2022sphere2vec} uses $\possamp^{\geotag}, \negsamp^{\negloc}$ 
    as positive and negative pairs.
    \item \textbf{SimCSE-based sampling ($\simcsetag$)} \cite{gao2021simcse} uses $\possamp^{\dropout}, \negsamp^{\dropout}$ as positive and negative pairs. Please refer to Appendix \ref{sec:simcse_describe} for a detailed description. 
\end{itemize}

Each corresponds to one loss component in our contrastive learning loss function by using either $\nce$ or $\mc$ objective shown in Equation \ref{equ:nce} and \ref{equ:mc}. So we define two versions of contrastive losses which both have three components.

\textit{The self-supervised binary ($\nce$) loss} $\lbi$ is defined as 
\begin{align}
\begin{split}
    \lbi(\Unlabelset)  &
    = \batchNLLloss(\Unlabelset) 
    + \neglocWeight \neglocNLLloss(\Unlabelset) 
    + \simcseWeight \simcseNLLloss(\Unlabelset) \\
    &
    = \lbi(\possamp^{\geotag}, \negsamp^{\batch}) 
    + \neglocWeight \lbi(\emptyset, \negsamp^{\negloc}) \\
    & + \simcseWeight \lbi(\possamp^{\dropout}, \negsamp^{\dropout})
\end{split}
\label{equ:NLLloss}
\end{align}
where $\neglocWeight$ and $\simcseWeight$ control the contribution of the last two loss components. Note here we use empty set as the positive pairs in $\neglocNLLloss(\Unlabelset)$ since $\possamp^{\geotag}$ has been considered in $\batchNLLloss(\Unlabelset) $.

\textit{The self-supervised multi-class ($\mc$) loss} $ \lmc$ is defined as 
\begin{align}
\begin{split}
    \lmc(\Unlabelset)  
    = &\batchContloss(\Unlabelset) 
    + \neglocContWeight \neglocContloss(\Unlabelset) 
    + \simcseContWeight \simcseContloss(\Unlabelset) \\
    = &\lmc(\possamp^{\geotag}, \negsamp^{\batch}, \batchTmp)
    + \neglocContWeight  \lmc(\possamp^{\geotag}, \negsamp^{\negloc},  \neglocTmp ) \\
    &+ \simcseContWeight \lmc(\possamp^{\dropout}, \negsamp^{\dropout},    \simcseTmp )
\end{split}
\label{equ:contloss}
\end{align}
where $\neglocContWeight$ and $\simcseContWeight$ are hyper-parameters. Although $\batchContloss(\Unlabelset)$ and $\neglocContloss(\Unlabelset) $ use the same positive pairs $\possamp^{\geotag}$, they are embedded in the Softmax function. So we need to use  $\possamp^{\geotag}$ in both loss components.

A naive contrastive pre-training for this dual-encoder architecture is to jointly training both encoders from scratch as CLIP \cite{radford2021clip} and ALIGN \cite{jia2021scaling} do for the image and text encoder.
However, from-scratch training will be problematics in \modelname.
Unlike CLIP and ALIGN's dual-encoder framework in which the text and image encoder have relatively the same number of trainable parameters,
the number of trainable parameters of the image encoder $\imgencprj()$ is 100 times larger than that of the location encoder $\enc()$. For example, the InceptionV3 image encoder we used for iNat2018 dataset has 41.8 million trainable parameters while the Space2Vec location encoder we used in both iNat2018 and fMoW dataset has only 0.4 million trainable parameters. Jointly training both encoders from scratch will yield overfitting issue for location encoder and underfitting issue for the image encoder.

Moreover, in text-image pre-training literature, LiT \cite{zhai2021lit} also reported that locking the image encoder during pre-training leads to a significant performance improvement. 
So we follow the practice of LiT \cite{zhai2021lit}, and utilize a pre-trained image network $\imgenc^{*}()$ and lock it during \modelfullname. The pre-trained image network $\imgenc^{*}()$ should not see the current image labels during pre-training stage. 
In other words, we first do image encoder pre-training as shown in the orange box of Figure \ref{fig:model-csp}. Then we lock $\imgencprj()$ and use it to pre-train $\enc()$ as shown in the red box of Figure \ref{fig:model-csp}. 
During \modelname, only the image projection layer $\imgprjw()$ is trained in the image encoder part.

\subsection{Supervised Fine-Turning} \label{sec:super_loss}
After \modelfullname, we follow the practice of \citet{chu2019geo,mac2019presence,mai2022sphere2vec} and fine-tune the image encoder $\imgencprj()$ and location encoder $\enc()$ separately on a small labeled dataset $\Labelset = \{(\th, \image, \classy)\}$ to test its performance in a few-shot learning setting.
The supervised fine-tuning stage corresponds to the green and blue box in Figure \ref{fig:model-csp}.
Their predictions are combined at the inference stage as \citet{mac2019presence,mai2020multiscale,mai2022sphere2vec} did.

\vspace{-0.1cm}
\paragraph{Image Encoder Fine Tuning}
We drop the projection layer $\imgprjw()$ and use a classification head $\imgcls()$ to 
process the image feature vector $\imgenc(\image)$ into logits over  image labels, i.e., $\imgcls(\imgenc(\image)) \in \Real^{\numclass}$. We fine-tune $\imgcls()$ with cross-entropy loss. $\numclass$ is the total number of classes.
This process corresponds to the green box in Figure \ref{fig:model-csp}. Please refer to Appendix \ref{sec:img_enc_img_enc_describe} for a detailed description of $\imgencprj()$ fine-tuning.

\paragraph{Location Encoder Fine Tuning}
As shown in the blue box of Figure \ref{fig:model-csp}, 
we use image labels in the training objective for location encoder fine tuning. 
Following
\citet{mac2019presence}, we used a 
\textit{presence-absence loss} function which converts the multi-class labels into
binary multi-labels.
A class embedding matrix $\classemb \in \Real^{\embdim \times \numclass}$ is used to supervisedly train the location encoder where $\classemb_{:, \classy} \in \Real^{\embdim}$ indicates the class embedding for the $\classy$th class. Given a set of training samples $\Labelset = \{(\th, \image, \classy)\}$ where $\classy$ indicates the class label, the loss function $\imgclsloss(\Labelset)$ is defined as:
\begin{align}
\begin{split}
 \imgclsloss(\Labelset)  = \beta  
     \lbi(\possamp^{\labelimg}, \emptyset)
     +  \lbi(\emptyset, \negsamp^{\labelimg}
     \cup \negsamp^{\labelloc})
\end{split}
\label{equ:imgloss}
\end{align}
Here $\lossweight$ is a hyperparameter for the weight of positive samples. The following positive and negative samples are used:
\begin{itemize}[leftmargin=*]
    \setlength\itemsep{0.3em}
    \item \textbf{Labeled positives}   $\possamp^{\labelimg} = \{(\enc(\th), \classemb_{:,\classy}) | (\th, \classy) \in \Labelset \}$.
    \item \textbf{Labeled negatives } $\negsamp^{\labelimg} = \{(\enc(\th), \classemb_{:, \classy_j}) | (\th, \classy) \in \Labelset , \classy_j \in \{1 .. \numclass\}\setminus \{\classy\} \}$.
    \item \textbf{Sampled negative locations}   $\negsamp^{\labelloc}  = \{( \enc(\th^{-}), \classemb_{:, \classy_j}) | (\th, \classy) \in \Labelset , \; \classy_j \in \{1 .. \numclass\} \}$, where $\th^{-}$ is a uniformly sampled locations from the surface of the sphere for each example $\th$.
\end{itemize}

\subsection{Model Inference} \label{sec:infer_method}

At inference time, we combined the predicted logits of fine-tuned $\enc()$ and  $\imgencprj()$ to give the final prediction as shown in the purple box of Figure \ref{fig:model-csp}. Given a location-image pair $(\th, \image)$, we estimate which category $\classy$ it belongs to by $P(\classy|\image,\th)$. 
According to \citet{mac2019presence}, if we assume $\image$ and $\th$ are conditionally independent given $\classy$, then based on Bayes' theorem, we have $P(\classy|\image,\th) \; \propto \; P(\classy|\th)P(\classy|\image)$. Here, $P(\classy|\image)$ can be estimated by the logits of $\imgcls(\imgenc(\image))$ at the $\classy$th class. For $P(\classy|\th)$, we have $P(\classy|\th) \; \propto \; \act(\enc(\th)\classemb_{:,\classy})$ where $\act(\cdot)$ is a sigmoid activation function.

 \vspace{-0.2cm}
\section{Experiments} \label{exp}
\vspace{-0.1cm}

In this work, we study the effectiveness of \modelname{} on two geo-aware image classification tasks - species fine-grained recognition and satellite image classification. 
We are particularly interested in how the dual-encode architecture performs in various \textit{few-shot learning} settings after \modelname.

For each task, three datasets are used to pre-train, fine-tune, and evaluate our \modelname~ models: $\Unlabelset_{train}$ is a set of unlabeled location-image pairs we use for pre-training;  
$\Labelset_{train}$ is a set of labeled location-image-class tuples we use for fine-tuning, where the size of $\Unlabelset_{train}$ is much larger than that of $\Labelset_{train}$, i.e., $|\Unlabelset_{train}| \gg |\Labelset_{train}|$; and
$\Labelset_{val}$ is a set of labeled location-image-class tuples we use for evaluation that can not be seen during fine-tuning.

\subsection{Models and Baselines} \label{sec:baseline}
In this work, we consider the following baselines: \setlist{nolistsep}
\begin{itemize}[leftmargin=*]
    \setlength\itemsep{0.2em}
\item \textbf{\imageonly} 
supervisedly fine-tune the image network $\imgcls(\imgenc())$ on the fine tuning dataset $\Labelset_{train}$ (See Figure \ref{fig:model-geo-ssl}). We use InceptionV3 \cite{szegedy2016rethinking} and ResNet50 \cite{ayush2020selfsup} as the image encoders on iNat2018 and fMoW respectively.
\item \textbf{\superonly} uses the dual-encoder architecture but is only supervisedly trained on $\Labelset_{train}$ (See Figure \ref{fig:model-geo-super}). We consider use $\aodha$ \cite{mac2019presence} and $\gridcell$ \cite{mai2020multiscale} as the location encoder which yield two models: \textbf{\superonly{}  (\aodha{})} and \textbf{\superonly{} (\gridcell)}.
\item \textbf{\mse} follows the same setup as \modelname{} (See Figure \ref{fig:model-csp}) except that during location encoder pre-training, 
it directly feeds the location embedding $\enc(\th)$ into a linear layer to regress the image feature vector $\imgenc(\image)$ with a Mean Square Error (MSE) loss. \mse{}  uses $\gridcell$ as the location encoder. 
\end{itemize}

We compare these baselines with different versions of \modelname. 
All \modelname~ models have the same training procedure, and use $\gridcell$ as their location encoders. The only difference is the contrastive loss function they use:
\setlist{nolistsep}
\begin{itemize}[leftmargin=*]
\setlength\itemsep{0.2em}
\item \textbf{\modelname-NCE-\inbatchtag\negloctag\simcsetag} uses the $\nce$ loss with all three loss components as shown in Equation \ref{equ:NLLloss}.

\item \textbf{\modelname-MC-\inbatchtag\negloctag\simcsetag} uses the $\mc$ loss with all three loss components as shown in Equation \ref{equ:contloss}.
\end{itemize}

\subsection{Fine-Grained Species Recognition}  \label{sec:sperecg_exp}  
We use the iNat2018 dataset\footnote{\url{https://github.com/visipedia/inat_comp/tree/master/2018}} \cite{van2018inaturalist} as a representative dataset to study the effectiveness of \modelname~ on species fine-grained recognition. iNat2018 is a large-scale species classification dataset with 8142 species categories. There are 437,513 training images of which 436,063 training images have geo-locations. On average each class has 53.6 training samples. We use all location-image pairs $\{(\th_i, \image_i)\}$ in iNat2018 training set as the unlabeled geo-tagged dataset $\Unlabelset_{train}$ for our \modelname{}. 
To create a few-shot learning task,
we perform a stratified sampling on the training dataset to select $\dataratio\%$ of training samples which constitute our few-shot supervised fine-tuning dataset $\Labelset_{train} = \{(\th, \image, \classy)\}$. 
The iNat2018 validation dataset is used for model evaluation to make our results comparable with previous work \cite{mac2019presence,mai2020multiscale,mai2022sphere2vec}. 
We use InceptionV3 network pre-trained on ImageNet as the image feature extractor $\imgenc^{*}()$ for iNat2018 dataset.

Table \ref{tab:unsuper_eval_in18} compares the Top1 accuracy of different training strategies on the iNat2018 validation dataset with different $\dataratio\%$. 
From Table \ref{tab:unsuper_eval_in18}, we can see that:
\setlist{nolistsep}
\begin{itemize}[leftmargin=*,noitemsep]
    \setlength\itemsep{0.3em}
    \item \imageonly{} (ImageNet) yields the lowest performances in all $\dataratio\%$ settings which indicates that considering location information is beneficial in all settings.
    
    \item \superonly{} (\gridcell{}) outperforms \superonly{} (\aodha{}) across all settings indicating that multi-scale location encoders (e.g., \gridcell) are effective for spatial distribution modeling. This confirms the results of \citet{mai2020multiscale}.
    
    \item Comparing the last three models, we can see the general patterns in all $\dataratio\%$ settings: \modelname-MC-\inbatchtag\negloctag\simcsetag > \modelname-NCE-\inbatchtag\negloctag\simcsetag > \mse. Since these three models only differ in terms of the location encoder pre-training strategies (the red box in Figure \ref{fig:model-csp}), this indicates that \modelname-MC-\inbatchtag\negloctag\simcsetag{} is the best location encoder pre-training objective.
    
    \item When $\dataratio\%=5\%, 10\%, 20\%$, compared with the \superonly{}, \modelname-MC-\inbatchtag\negloctag\simcsetag{} have relative performance improvements of 10.4\%, 34.3\%, and 16.6\% which indicates the effectiveness of \modelfullname.
    
    \item When $\dataratio\%=100\%$, \modelname-MC-\inbatchtag\negloctag\simcsetag{} still yields better results than \superonly{} (\gridcell). This indicates that our $\modelname$ is beneficial even in a fully supervised setting.
\end{itemize}

To understand the effectiveness of each loss component in \modelname{} (see Figure \ref{fig:csp_loss} and Equation \ref{equ:contloss}), we conduct an ablation study on the iNat2018 dataset  with different $\dataratio$ and report the results in Table \ref{tab:unsuper_eval_in18_ablation}. We can see that each component contributes to the final model performance. Deleting any of them will lead to performance drops.

To understand the effect of location embedding dimension $\locdim$ on the model performance, we conduct an additional ablation study of $\locdim$ on the iNat2018 dataset  with different $\dataratio$ and report the results in Table \ref{tab:unsuper_eval_in18_ablation_locdim}. We can see that at the few-shot setting $\dataratio\%=5\%, 10\%, 20\%$, models with $\locdim = 256$ achieve the best performance. In the fully supervised setting, the model with $\locdim = 1024$ leads to the best performance.

Last but not least, we also explore whether our \modelname{} is effective on different image encoders. We conduct an ablation study of different $\imgenc()$ on the iNat2018 dataset  with $\dataratio\%=5\%$. 
Table \ref{tab:unsuper_eval_in18_ablation_imgenc} summarizes the results. We can see that no matter which $\imgenc()$ we use, Inception V3 or ViT, our \modelname-MC-\inbatchtag\negloctag\simcsetag{} consistently yields the best results, and ViT improves the model performance a lot.

\begin{figure*}[ht!]
	\centering \vspace*{-0.2cm}
	\begin{subfigure}[b]{0.24\textwidth}  
		\centering 
		\includegraphics[width=\textwidth]{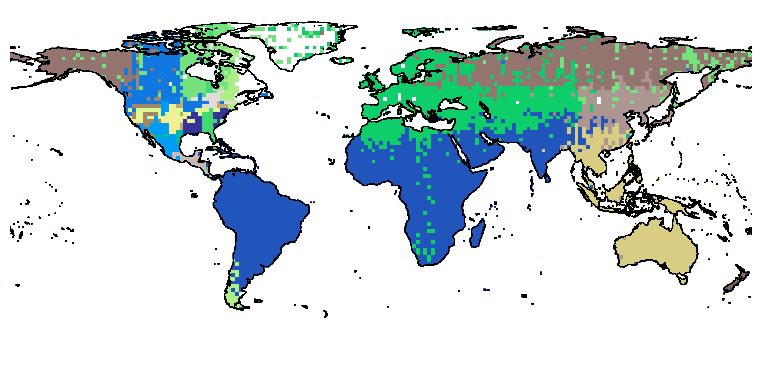}
		\vspace*{-0.9cm}\caption[]{{ MC Unsupervised	}}    
		\label{fig:inat18_mc_unsup}
	\end{subfigure}
\begin{subfigure}[b]{0.24\textwidth}  
		\centering 
		\includegraphics[width=\textwidth]{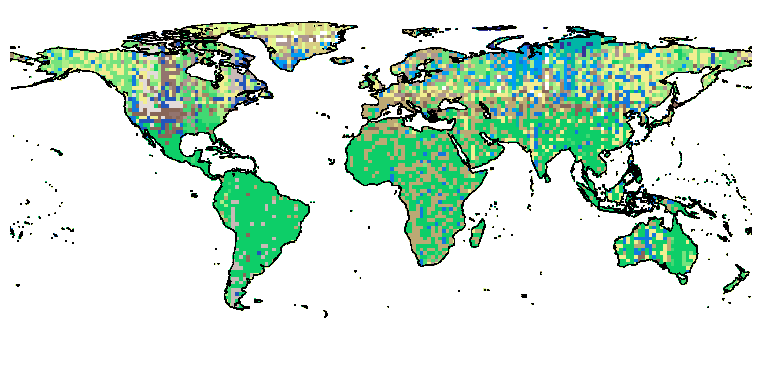}
		\vspace*{-0.9cm}\caption[]{{ MC Supervised}}    
		\label{fig:inat18_mc_sup}
	\end{subfigure}
\begin{subfigure}[b]{0.24\textwidth}  
		\centering 
		\includegraphics[width=\textwidth]{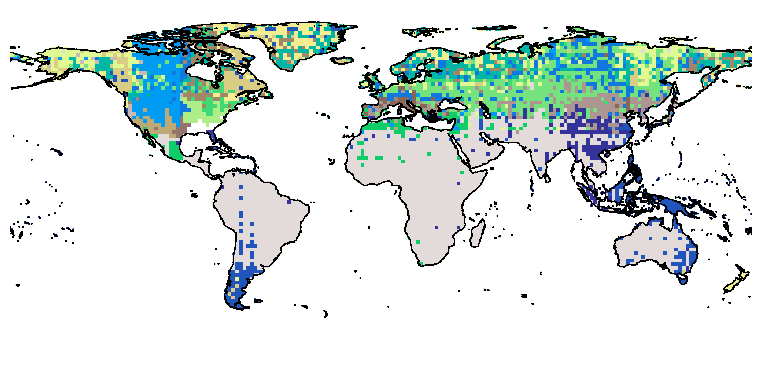}
		\vspace*{-0.9cm}\caption[]{{ NCE Unsupervised	}}    
		\label{fig:inat18_nce_unsup}
	\end{subfigure}
\begin{subfigure}[b]{0.25\textwidth}  
		\centering 
		\includegraphics[width=\textwidth]{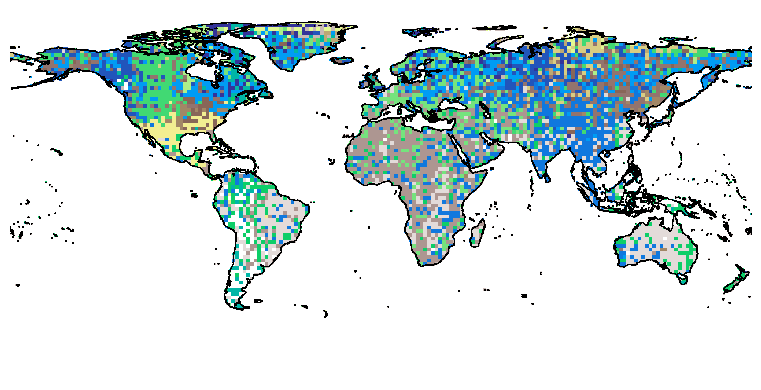}
		\vspace*{-0.9cm}\caption[]{{ NCE Supervised	}}   
		\label{fig:inat18_nce_sup}
	\end{subfigure}
	\vspace{-0.3cm}
	\caption{
	Location embedding before  and after supervised fine-tuning for iNat2018.
	} 
	\label{fig:inat18_clustering}
\end{figure*}

To investigate how well \modelname~ learns location representation, 
we sample a set of regular grid points all over the world and compute their embeddings with
the location encoder $\enc()$. 
The resulting location embeddings are hierarchically clustered. 
The results are shown in Figure \ref{fig:inat18_clustering}. 
Figure \ref{fig:inat18_mc_unsup} and \ref{fig:inat18_nce_unsup} show the clustering results after \textit{\modelname-MC-\inbatchtag\negloctag\simcsetag} or \textit{\modelname-NCE-\inbatchtag\negloctag\simcsetag} pre-training, 
while Figure \ref{fig:inat18_mc_sup} and \ref{fig:inat18_nce_sup} show the clustering results after supervised fine-tuning on respective models.
Some interesting clustering patterns merge in Figure \ref{fig:inat18_mc_unsup} and \ref{fig:inat18_nce_unsup}. 
For example, the clustering patterns in Figure \ref{fig:inat18_mc_unsup} show some regional effects that are somewhat similar to the Köppen climate classification\footnote{\url{https://en.wikipedia.org/wiki/K\%C3\%B6ppen_climate_classification}}. 
This makes sense since the pre-training with location-image pairs is learning  the spatial distribution of species and their environment,
which is highly related to climate zones.
The clusters in the US are smaller since the iNat2018 training dataset has much more data in the US
(See Figure \ref{fig:inat18_train_locs} in Appendix \ref{sec:inat18_data_stat}).

\begin{table}[t!]
\caption{
The Top1 accuracy of different models and training strategies on the iNat2018 validation dataset for the species fine-grain recognition task with different training data ratios, where $\dataratio\% = 100\%$ indicates the fully supervised setting. We run each model 5 times and report the standard deviation in ``()''. 
}
\label{tab:unsuper_eval_in18}
\centering \setlength{\tabcolsep}{3pt}
\vspace{-0.3cm}
{\scriptsize
\begin{tabular}{c|c|c|c|c}
\toprule
Ratio $\dataratio\%$                  & 5\%                  & 10\%                  & 20\%                  & 100\%                 \\ \hline
\makecell{\imageonly{} (ImageNet)\\ \cite{szegedy2016rethinking}} & 5.28 (-)             & 12.44 (-)             & 25.33 (-)             & 60.2 (-)              \\ \hline
\makecell{\superonly{} (\aodha)  \\ \cite{mac2019presence}}     & 7.12 (0.02)          & 12.50 (0.02)          & 25.36 (0.03)          & 72.41 (-)             \\
\makecell{\superonly{} (\gridcell) \\ \cite{mai2020multiscale}}     & 8.16 (0.01)          & 14.65 (0.03)          & 25.40 (0.05)          & 72.98 (0.04)          \\ \hline
\mse                   & 8.15 (0.02)          & 17.80 (0.05)          & 27.56 (0.02)          & 73.27 (0.02)          \\ \hline
\modelname-NCE-\inbatchtag\negloctag\simcsetag           & 8.65 (0.02)          & 18.75 (0.12)          & 28.15 (0.07)          & 73.33 (0.01)          \\
\modelname-MC-\inbatchtag\negloctag\simcsetag            & \textbf{9.01 (0.02)} & \textbf{19.68 (0.05)} & \textbf{29.61 (0.03)} & \textbf{73.79 (0.02)} \\ \bottomrule
\end{tabular}
}
\vspace{-0.3cm}
\end{table}
\vspace{-0.5cm}
\begin{table}[t!]
\caption{
Ablation studies on different \modelname-MC-* pretraining objectives on the iNat2018 validation dataset with different $\dataratio\%$. Here, \modelname-MC-\inbatchtag\negloctag\simcsetag{} indicates the \modelname{} training on the $\mc$ loss with all three components. \modelname-MC-\inbatchtag\negloctag{} deletes the SimCSE $\simcseContloss(\Unlabelset)$ component in Equation \ref{equ:contloss}. The rest models follow similar logic.
}
\label{tab:unsuper_eval_in18_ablation}
\centering \small
\vspace{-0.3cm}
{\begin{tabular}{l|c|c|c|c}
\toprule
Ratio $\dataratio\%$                  & 5\%                  & 10\%                  & 20\%                  & 100\%                 \\ \hline
\modelname-MC-\inbatchtag\negloctag\simcsetag & \textbf{9.01} & \textbf{19.68} & \textbf{29.61} & \textbf{73.79} \\ \hline
\modelname-MC-\inbatchtag\simcsetag  & 8.63          & 19.60           & 29.52          & 73.15          \\
\modelname-MC-\inbatchtag\negloctag  & 8.40           & 17.17          & 26.63          & 73.36          \\
\modelname-MC-\inbatchtag   & 8.16          & 16.58          & 25.89          & 73.10 \\
\bottomrule
\end{tabular}
}

\vspace{-0.3cm}
\end{table}
\begin{table}[t!]
\caption{
Ablation studies on different location embedding dimensions $\locdim$ on the iNat2018 validation dataset with different $\dataratio\%$. 
}
\label{tab:unsuper_eval_in18_ablation_locdim}
\centering \small
\vspace{-0.3cm}
{\begin{tabular}{c|c|c|c|c|c}
\toprule
           & $\locdim$    & 5\%           & 10\%           & 20\%           & 100\%          \\ \hline
\modelname-MC-\inbatchtag\negloctag\simcsetag & 64   & 7.64          & 16.57          & 25.31          & 71.76          \\
\modelname-MC-\inbatchtag\negloctag\simcsetag & 128  & 8.5           & 19.35          & 29.11          & 72.89          \\
\modelname-MC-\inbatchtag\negloctag\simcsetag & 256  & \textbf{9.01} & \textbf{19.68} & \textbf{29.61} & 73.62          \\
\modelname-MC-\inbatchtag\negloctag\simcsetag & 512  & 8.97          & 18.8           & 27.96          & 73.67          \\
\modelname-MC-\inbatchtag\negloctag\simcsetag & 1024 & 8.78          & 17.94          & 26.65          & \textbf{73.79} \\ \bottomrule
\end{tabular}
}
\vspace{-0.2cm}
\end{table}

\vspace{-0.9cm}
\begin{table}[t!]
\caption{
Ablation studies on different image neural network $\imgenc()$   (InceptionV3  \citep{szegedy2016rethinking} and ViT  \citep{dosovitskiy2021vit})
on the iNat2018 validation dataset with $\dataratio\%=5\%$. 
}
\label{tab:unsuper_eval_in18_ablation_imgenc}
\centering \small
\vspace{-0.3cm}
{\footnotesize \begin{tabular}{c|c|c}
\toprule
$\imgenc()$             & \makecell{Inception V3 }     & \makecell{ViT 
}            \\ \hline
\makecell{\imageonly{} (ImageNet) \\ \cite{szegedy2016rethinking}} & 5.28          & 12.46          \\ \hline
\makecell{\superonly{} (\aodha)  \\ \cite{mac2019presence}}     & 7.12          & 18.66          \\
\makecell{\superonly{} (\gridcell) \\ \cite{mai2020multiscale}}     & 8.16          & 18.68          \\ \hline
\mse                   & 8.15          & 20.02          \\ \hline
\modelname-NCE-\inbatchtag\negloctag\simcsetag           & 8.65          & 20.16          \\
\modelname-MC-\inbatchtag\negloctag\simcsetag            & \textbf{9.01} & \textbf{20.78} \\ \bottomrule
\end{tabular}
}
\vspace{-0.7cm}
\end{table} \bigskip \bigskip

\begin{table}[ht!]
\caption{
The Top1 accuracy of different models and training strategies on the fMoW val dataset  for the satellite image classification task with different training data ratios, where $\dataratio\% = 100\%$ indicates fully supervised setting.
We report the standard errors (SE) over 5 different runs. 
	}
	\vspace{-0.3cm}
	\label{tab:unsuper_eval_fmow}
	\centering \small
	\setlength{\tabcolsep}{2pt}
{\scriptsize
\begin{tabular}{c|c|c|c|c}
\toprule
Ratio $\dataratio\%$             & 5\%                   & 10\%                 & 20\%                  & 100\%              \\ \hline
\makecell{\imageonly{} (Tile2Vec) \\ \cite{jean2019tile2vec} } & 59.41 (0.23)          & 61.91 (0.31)          & 62.96 (0.51)          & 64.45 (0.37)       \\
\makecell{\imageonly{} (Geo-SSL) \\ \cite{ayush2020selfsup} }  & 65.22 (-)             & 66.46 (-)             & 67.66 (-)             & 69.83 (-)          \\ \hline
\makecell{\superonly{} (\aodha{}) \\ \cite{mac2019presence}}     & 66.67 (0.03)          & 68.22 (0.01)          & 69.45 (0.01)          & 70.30 (0.02)        \\
\makecell{\superonly{} (\gridcell{}) \\\cite{mai2020multiscale}}     & 67.01 (0.02)          & 68.91 (0.04)          & 70.20 (0.03)           & 70.77 (0.03)       \\ \hline
\mse                   & 67.06 (0.04)          & 68.90 (0.05)           & 70.16 (0.02)          & 70.45 (0.01)       \\ \hline
\modelname-NCE-\inbatchtag\negloctag\simcsetag           & 67.29 (0.03)          & 69.20 (0.03)           & 70.65 (0.02)          & 70.89 (0.04)       \\
\modelname-MC-\inbatchtag\negloctag\simcsetag            & \textbf{67.47 (0.02)} & \textbf{69.23 (0.03)} & \textbf{70.66 (0.03)} & \textbf{71.00 (0.02)} \\
\bottomrule
\end{tabular}
}
\vspace{-0.3cm}
\end{table}

\begin{figure*}[ht!]
	\centering \vspace*{-0.2cm}
	\begin{subfigure}[b]{0.24\textwidth}  
		\centering 
		\includegraphics[width=\textwidth]{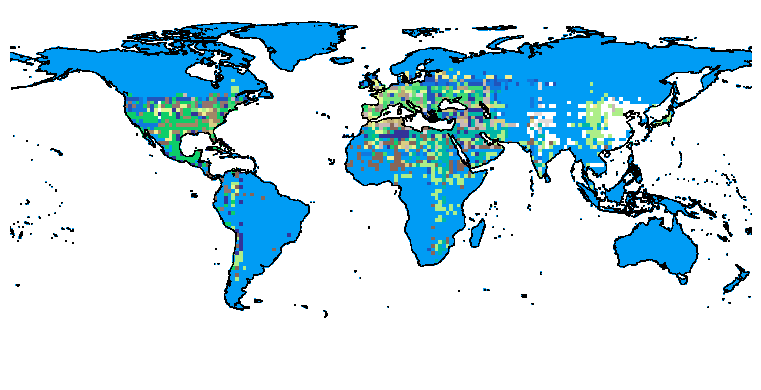}
		\vspace*{-0.9cm}\caption[]{{ MC Unsupervised	}}    
		\label{fig:fmow_mc_unsup}
	\end{subfigure}
	\hfill
	\begin{subfigure}[b]{0.24\textwidth}  
		\centering 
		\includegraphics[width=\textwidth]{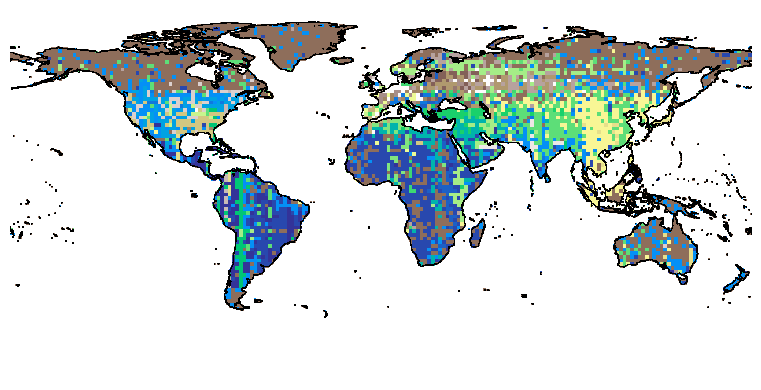}
		\vspace*{-0.9cm}\caption[]{{ MC Supervised}}    
		\label{fig:fmow_mc_sup}
	\end{subfigure}
	\hfill
	\begin{subfigure}[b]{0.24\textwidth}  
		\centering 
		\includegraphics[width=\textwidth]{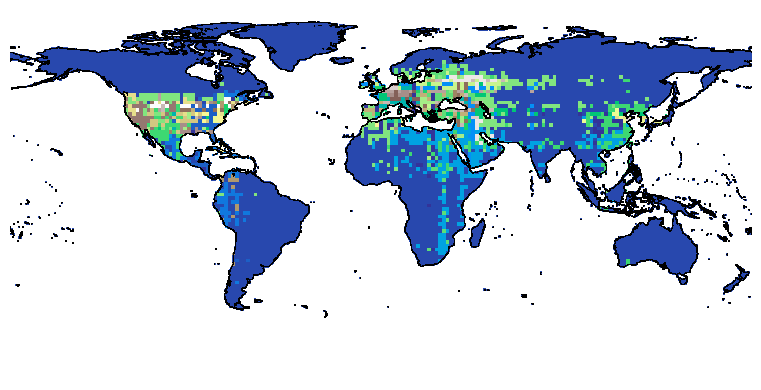}
		\vspace*{-0.9cm}\caption[]{{ NCE Unsupervised	}}    
		\label{fig:fmow_nce_unsup}
	\end{subfigure}
	\hfill
	\begin{subfigure}[b]{0.24\textwidth}  
		\centering 
		\includegraphics[width=\textwidth]{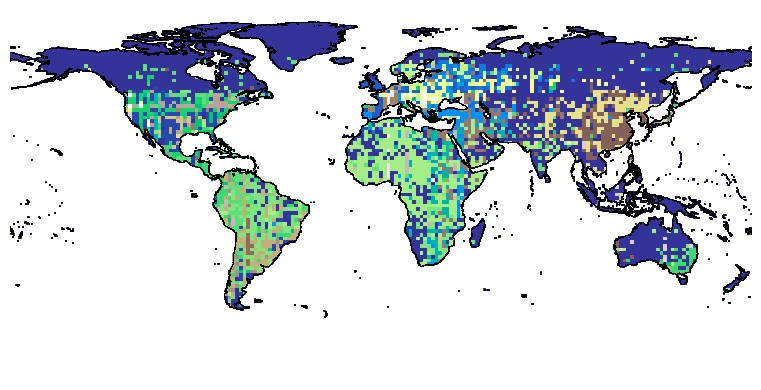}
		\vspace*{-0.9cm}\caption[]{{ NCE Supervised	}}   
		\label{fig:fmow_nce_sup}
	\end{subfigure}
	\vspace{-0.3cm}
	\caption{
	Location embedding before and after supervised fine tuning for fMOW.
	} 
	\label{fig:fmow_clustering}
	\vspace{-0.3cm}
\end{figure*}

\vspace{10pt}
\subsection{Satellite Image Classification}  \label{sec:satellite_exp} 

A similar procedure is carried out on fMoW\footnote{\url{https://github.com/fMoW/dataset}} dataset \cite{christie2018functional}, which has 62 different geospatial object classes, and
363,570  location-image pairs. 
We use all location-image pairs as $\Unlabelset_{train}$, and stratified sample $\dataratio\%$ labeled location-image pairs from the training dataset as $\Labelset_{train}$. 
We use similar training, and evaluation protocol as Section \ref{sec:sperecg_exp}. The ResNet50 checkpoint after Geo-SSL's MoCo-V2+TP self-supervised pre-training on unlabeled fMoW dataset \cite{ayush2020selfsup} is used as the pre-trained image feature extractor $\imgenc^{*}()$ for all models. 

Table \ref{tab:unsuper_eval_fmow} compares the evaluation results (Top1 accuracy) among different models and training strategies on the fMoW val dataset after fine-tuning on $\dataratio\%$ fMoW training samples where $\dataratio\% \in \{5\%, 10\%, 20\%, 100\%\}$. 
We can see that Table \ref{tab:unsuper_eval_fmow} shows similar patterns as those of Table \ref{tab:unsuper_eval_in18}: 
\begin{itemize}[leftmargin=*]
    \setlength\itemsep{0.5em}
    \item \imageonly{} (Geo-SSL) yields  better results than \imageonly{} (Tile2Vec) across different $\dataratio\%$. But both \imageonly{} models still give the lowest performance than all other settings with all $\dataratio\%$. This confirms the importance of jointly learning location representations. However, \imageonly{} (Geo-SSL) gives a relatively good performance (65.22\%) even when $\dataratio\% = 5\%$.
    That is because we use the Geo-SSL's MoCo-V2+TP checkpoint which is directly pre-trained on the unlabeled fMoW training dataset. In contrast, in Table \ref{tab:unsuper_eval_in18}, \imageonly{} used an InceptionV3 model pre-trained on ImageNet, not on the iNat2018 training dataset.
    
    \item Similar to the results in Table \ref{tab:unsuper_eval_in18}, \superonly{} (\gridcell) outperforms \superonly{} (\aodha) in all settings which shows the effectiveness of \gridcell{} over \aodha{}.
    
    \item \modelname-MC-\inbatchtag\negloctag\simcsetag{} outperforms all models and yields super or comparable results of \modelname-NCE-\inbatchtag\negloctag\simcsetag. However, the margins are relatively small compared with those of Table \ref{tab:unsuper_eval_in18}. The performance improvements mainly come from the location encoder's ability to do spatial distribution modeling. Compared with species distribution, the geographic distributions of land use types are very complex and hard to differentiate from each other. For example, factories and multi-unit residential buildings are both man-made geographic entities. Both their distributions are correlated with population distributions and are hard to differentiate. Moreover, sometimes they also show similar appearance in remote sensing images. So it is rather hard to use a location encoder to differentiate one land use type from the other based on their geographic distribution. 
We think a more powerful location encoding is needed to differentiate them. But this is beyond the scope of this paper.

\end{itemize}

Similar to the iNat2018 dataset, in the fMoW dataset, the embedding clustering results of the pre-trained and fine-tuned location encoders 
are visualized in Figure \ref{fig:fmow_clustering}. 
We can see that more fine-grained clusters are generated in the US after \textit{\modelname-MC-\inbatchtag\negloctag\simcsetag{}}/\textit{\modelname-NCE-\inbatchtag\negloctag\simcsetag{}} pre-training, while the representation is updated to be more detailed after location encoder fine-tuning. Compared with Figure \ref{fig:inat18_clustering}, the regional effect is less clear which also shows the difficulty to model the spatial distributions of land use types.

 \section{Conclusion and Discussion}  \label{sec:conclusion}

In this work, we proposed \modelfullname~(\modelname), a self-supervised framework to learn the alignment between locations and images based on large unlabeled geo-tagged images. 
Similar to recent popular image-text pre-training models such as CLIP and ALIGN, \modelname~ utilizes a dual-encoder architecture to separately encode the location and image. The resulting location and image representation are contrasted against each other to form a contrastive pre-training objective. To validate the effectiveness of \modelname, we conduct experiments on two geo-aware image classification tasks: species fine-grained recognition on iNat2018 dataset and satellite image classification on the fMoW dataset. 
Experiments results show that \modelname{} can improve model performance on both datasets under different labeled training data sampling ratios.
On the iNat2018 dataset \modelname{} can significantly boost the model performance with 10-34\% relative improvement in several few-shot settings ( $\dataratio\%=\{5\%, 10\%, 20\%\}$) and still be able to improve model performance when $\dataratio=100\%$.

To the best of our knowledge, our work is the first one to show the great potential of learning the geospatial-visual alignment for model pre-training. Although we only investigate the effectiveness of our \modelname~framework on location-image pre-training in this work, \modelname~ can be easily extended to learn the alignment between location (or time) and data in other modalities such as text for different downstream tasks such as geo-aware text classification. 
We put this as one of our future works. 
Moreover, in this work, we only use the existing geo-tagged datasets (e.g., iNat2018 and fMoW) as a proxy for unlabeled location-image pairs. In the future, we would like to construct larger-scale unlabeled geo-tagged image datasets based on publicly available satellite images with which we expect to see a larger performance improvement. 
In this work, we only use single geo-coordinates for geospatial-visual contrastive representation learning. In the future, we can explore more complex geometries such as polylines \citep{xu2018encoding} and polygons \citep{mai2023towards}. The proposed \modelname~ framework can be seen as a step towards the multimodal foundation models for geospatial artificial intelligence \citep{mai2022towards,mai2023opportunities}.

 \section{Ethics Statements}

Our code and used datasets are available from \url{https://gengchenmai.github.io/csp-website/}. 
We do not find any negative societal impact of our research.  \section*{Acknowledgements} \label{sec:ack}

This research is supported in part by 
ODNI, IARPA (2021-2011000004),  NSF (\#1651565), CZ Biohub, and Stanford HAI. 
The views and conclusions contained herein are those of the authors and should not be interpreted as necessarily representing the official policies, either expressed or implied, of ODNI, IARPA, or the U.S. Government. The U.S. Government is authorized to reproduce and distribute reprints for governmental purposes not-withstanding any copyright annotation therein.


\bibliographystyle{icml2023}

\newpage
\appendix
\onecolumn

\newpage
\section{Appendix}

\subsection{A Detailed Version of Figure \ref{fig:model-geo-ssl}} \label{sec:mode-geo-ssl-fig}

\begin{figure*}[ht!]
	\centering \tiny
\includegraphics[width=0.95\textwidth]{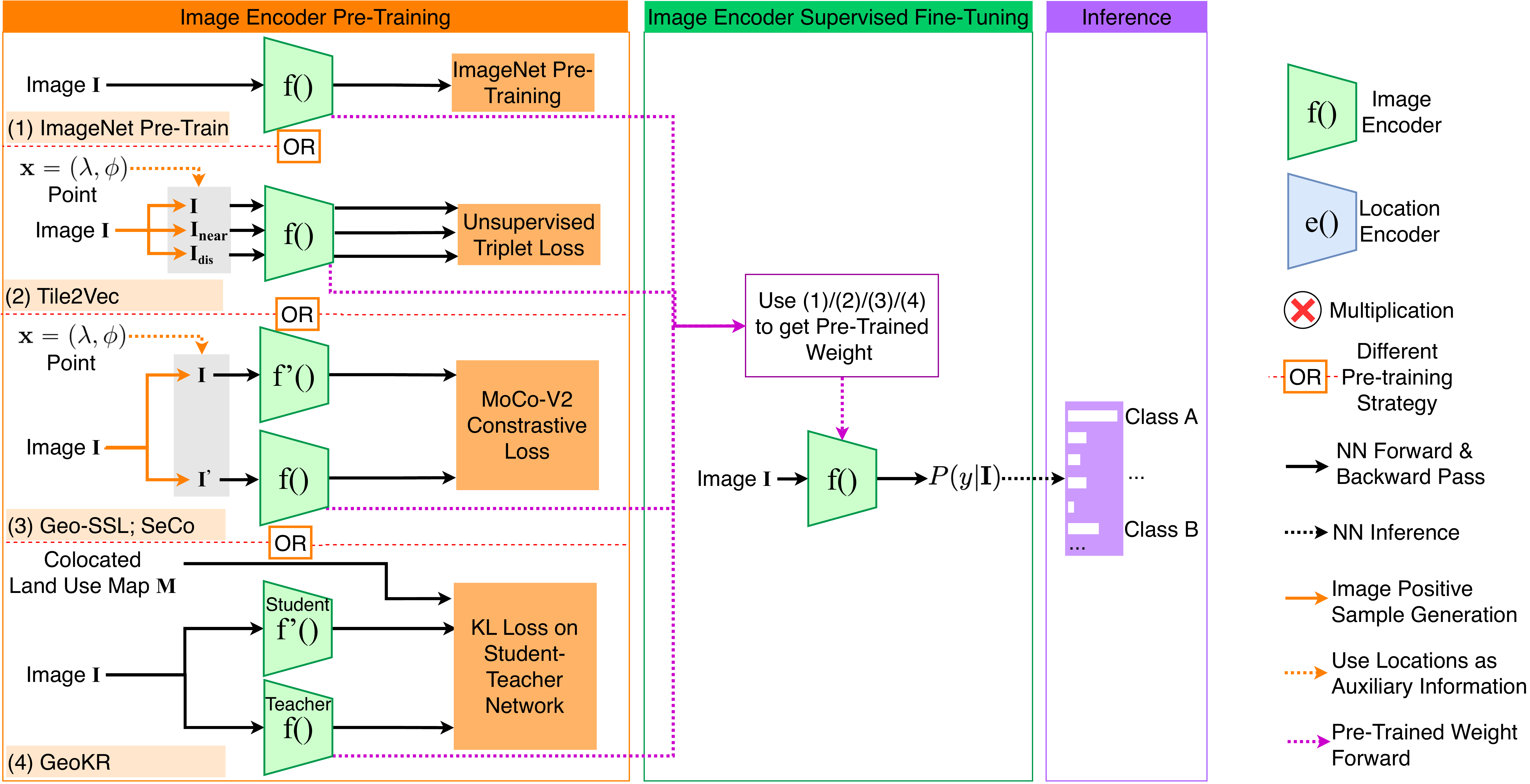}\vspace*{-0.2cm}
	\caption{
	A detailed version of Figure \ref{fig:model-geo-ssl} to show four different ways to pretrain image encoder $\imgencprj()$ (orange box): 
	(1) \textbf{ImageNet Pretraining} \cite{deng2009imagenet}: pre-training $\imgencprj()$ on ImageNet dataset; 
	(2) \textbf{Tile2Vec} \cite{jean2019tile2vec}: pretraining $\imgencprj()$ with an unsupervised triplet loss such that the embeddings of spatially nearby image tiles are more similar than those of distant tiles.
	(3) \textbf{Geo-SSL} \cite{ayush2020selfsup} and SeCo\cite{manas2021seasonal}: pretraining $\imgencprj()$ with a Momentum Contrast (MoCo-v2) \cite{chen2020mocov2} style constrastive loss in which they used locations as auxiliary information to generate spatially aligned (remote sensing) images at different timestamps as positive samples; 
	(4) \textbf{GeoKR} \cite{li2021geographical}: pretraining $\imgencprj()$ in a teacher-student network by minimizing the KL (Kullback–Leibler) loss between the image representations and a spatially aligned auxiliary data such as land cover maps $\landcovermap$. 
	The pre-trained weights of $\imgencprj()$ are fine-tuned in a supervised manner (green box) for image classification. Here, location is only used as auxiliary information for image encoder pre-training while being ignored during supervised learning stage. } 
	\label{fig:model-geo-ssl-detail}
	\vspace*{-0.15cm}
\end{figure*}

 \subsection{Model Architecture Training Detail
} \label{sec:model_train_detail}

\subsubsection{Training Details of Image Encoder $\imgencprj()$}
\label{sec:img_enc_img_enc_describe}

\begin{figure*}[ht!]
	\centering \tiny
\includegraphics[width=0.8\textwidth]{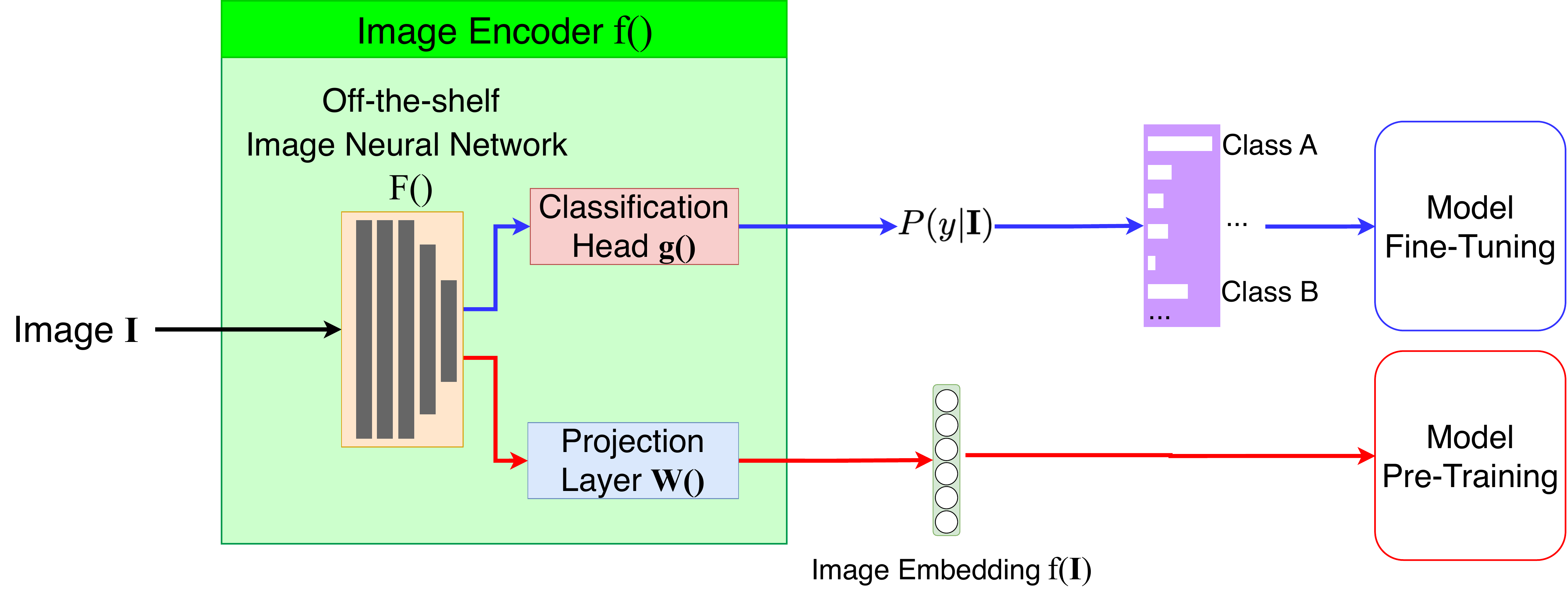}\vspace*{-0.2cm}
	\caption{
	A detailed illustration of the image encoder $\imgencprj()$ we use in Figure \ref{fig:model_illus} for model pre-training and fine-tuning. 
	} 
	\label{fig:img_enc}
	\vspace*{-0.15cm}
\end{figure*}

Figure \ref{fig:img_enc} is a detailed illustration of how we use the image encoder $\imgencprj()$ for model pre-training and fine-tuning. On both iNat2018 and fMoW dataset, we use off-the-shelf image neural network $\imgenc()$ to first encode the given image $\image$ into a $\imgembdim$ dimension image feature vector $\imgenc(\image) \in \Real^{\imgembdim}$.
On the iNatlist dataset, two pre-trained image models are used -- 1)  ImageNet pre-trained InceptionV3 \cite{szegedy2016rethinking} from PyTorchVision library\footnote{\url{https://pytorch.org/vision/main/models/generated/torchvision.models.inception\_v3.html\#torchvision.models.inception\_v3}} and 2) ImageNet pre-trained Vision Transformer\footnote{More specifically, we use the \textit{vit\_tiny\_patch16\_224} implementation from Huggingface timm at \url{https://github.com/huggingface/pytorch-image-models/blob/main/timm/models/vision_transformer.py} by following \citet{cong2022satmae}.} (ViT) \citep{dosovitskiy2021vit} from Huggingface timm library. See Table \ref{tab:unsuper_eval_in18_ablation_imgenc} for the performance comparison of these two models.
On the fMoW dataset, we use Geo-SSL \cite{ayush2020selfsup} pretrained ResNet50 \cite{he2015resnet} as $\imgenc()$.

During the \modelname~ pre-training stage, we feed  $\imgenc(\image)$ into a projection layer $\imgprjw()$ which projects the image feature $\imgenc(\image_i) \in \Real^{\imgembdim}$ into $\embdim$ dimension such that a contrastive learning objective can be formed between $\enc(\th_i)$ and $\imgencprj(\image_i)$. This illustrates as red arrows in Figure \ref{fig:img_enc}. 

During the image encoder fine-tuning stage, we drop the projection layer $\imgprjw()$ and append a classification head $\imgcls()$ to the end of $\imgenc()$ which maps the image feature vector $\imgenc(\image)$ into logits over each image label, i.e., $\imgcls(\imgenc(\image)) \in \Real^{\numclass}$. $\numclass$ is the total number of classes. This illustrates as blue arrows in Figure \ref{fig:img_enc}. Here, both $\imgprjw()$ and $\imgcls()$ are implemented as multi-layer perceptrons. 
In practice, on the iNat2018 dataset, we fine-tune the whole image encoder $\imgcls(\imgenc(\image))$ instead of only linear probing the $\imgcls()$ because 1) both used Inception V3 and ViT image neural network are previously pre-trained on ImageNet dataset whose images are different from the images in iNat2018; 2) Empirical experiments show that fine-tuning the whole architecture yields better performances; 3) The same practice was adopted by \citet{mac2019presence,mai2020multiscale}. 
On the fMoW dataset, we use Geo-SSL \cite{ayush2020selfsup} pretrained ResNet50 \cite{he2015resnet} and only linear probe on the $\imgcls()$. That is because the used $\imgenc()$ is self-supervised pre-trained on the same dataset.

\subsubsection{Training Details of Location Encoder $\enc()$}
\label{sec:loc_enc_detail_describe}

In terms of the location encoder, 
we use the Space2Vec \gridcell{} \cite{mai2020multiscale} as the location encoder for both datasets except for \imageonly{} and \superonly{} (\aodha{}) model. \imageonly{} does not use location encoders and \superonly{} (\aodha{}) uses \aodha{} location encoder.

During model pre-training, after pre-training the image encoder $\imgencprj()$, we lock $\imgencprj()$ and use it to pre-train $\enc()$ as shown in the red box of Figure \ref{fig:model-csp}. 
Location encoder fine-tuning details have been described in Section \ref{sec:super_loss}

\subsubsection{Model Implementation Details}
\label{sec:imple_detail_describe}
All models are implemented in PyTorch and trained on a Linux machine with 252GB memory and two GeoForce CUDA cores. The code, data, and pre-trained models of this work are all available at \url{https://gengchenmai.github.io/csp-website/}. 
\subsection{Implementation Details of SimCSE}
\label{sec:simcse_describe}

The implementation of SimCSE shown in Figure \ref{fig:simcse_loss} is inspired by \citet{gao2021simcse}. Basically, we have initialized two location encoder towers with identical structures. They share the parameters but they use independently sampled dropout masks. The two dropout masks are sampled independently for every training examples during \modelname{} pretraining. The masks are automatically generated by the dropout layers.

In the implementation, we \textbf{simply feed the same location $\th_i$ to the same location encoder twice and get two location embeddings $\enc(\th_i)$ and $\enc'(\th_i)$. Since they are based on two separate forward passes, they are based on different dropout masks}. When we obtain $\enc'(\th_i)$, we not only get the location embedding for $\th_i$ but also get embeddings for all locations in the same mini-batch. Among all these locations in the mini-batch, we select the embedding of location $ \th_j$ – $\enc'(\th_j)$ where $j \neq i$. Since $\enc'(\th_j)$ and $\enc'(\th_i)$ are generated based on the same forward pass, they share the same dropout mask. So the pair $(\enc(\th_i), \enc'(\th_i))$ is the dropout positive sample (the only difference is the dropout mask) and $(\enc(\th_i), \enc'(\th_j))$ is the negative pair who use different dropout masks and encode different input locations. In short, SimCSE simply uses dropout as a data augmentation tool to generate positive samples.

\subsection{Model Hyperparameter Tuning}  \label{sec:hyper_search}
Since the self-supervised learning takes very long time to tune and hard to evaluate,
we first perform a grid search 
to tune the hyperparameters related to supervised fine tuning stage for the location encoder without self-supervised pre-training. 
In other words, we tune those hyperparameters on the \superonly{} model. The best hyerparameter combination for \superonly{} is used for all \textit{\modelname} models and \textit{\mse}. 
The major hyperparameters we tune include the fine-tuning learning rate $\lr = [0.01, 0.005, 0.002, 0.001,  0.0005, 0.00005]$, the \gridcell's minimum scaling factor $\minscale = [0.1, 0.01, 0.001, 0.0005, 0.0001]$, as well as the hyperparameters of location encoder's multi-layer perceptron $\peffn(\cdot)$ such as its activation function $\sigmoid_{\enc} = [ReLU,  LeakyReLU, GELU]$, the number of hidden layers $\numresnet = [1,2,3]$,  the number of neurons  $\numneuron = [256, 512, 1024]$, and the dropout rate in $\peffn(\cdot)$  $\dropout = [0.1, 0.2, 0.3, 0.4, 0.5, 0.6, 0.7]$. The hyperparameters are tuned one-by-one sequentially by following the order: $\lr$, $\minscale$, $\sigmoid_{\enc}$, $\numresnet$, $\numneuron$, and $\dropout$.
Based on the experiment, the best hyperparameter combination for the few-shot learning on iNat2018 dataset is $\lr=0.0005$, $\minscale=0.01$, $\sigmoid_{\enc}=LeakyReLU$, $\numresnet=1$, $\numneuron=512$, $dropout=0.5$.
As for the few-shot learning on fMoW dataset, the best hyperparameter combination is $\lr=0.001$, $\minscale=0.01$, $\sigmoid_{\enc}=GELU$, $\numresnet=1$, $\numneuron=512$, $dropout=0.5$. Based on hyperparameter tuning results, we find out that a deeper $\peffn(\cdot)$, a larger $\numresnet$, for the location encoder does not necessarily increase the model performance.

After we get the best hyperparameter for the location encoder, we fix them and do a grid search to find the best hyperparameters for self-supervised pre-training. The main hyperparameter which we tune is
the self-supervised training learning rate $\lrunsuper = [0.01, 0.001, 0.0005, 0.0002, 0.0001,0.00001, 0.000002]$.
For \textit{\modelname-MC-*}, we tune
the negative location loss weight $\neglocContWeight$, SimCSE loss weight $\simcseContWeight$, the number of sampled negative locations $\neglocsize$, three temperatures $\batchTmp$, $\neglocTmp$, and $\simcseTmp$. 
For \textit{\modelname-NCE-*}, we also fine tune $\neglocWeight$ and $\simcseWeight$.

On the iNat2018 dataset, the best hyperparameter for \textit{\mse} is  $\lrunsuper =0.000002$. For \textit{\modelname-MC-*}, the best hyperparameter combination is $\lrunsuper =0.0002$, $\neglocContWeight=1$, $\simcseContWeight=1$, $\neglocsize=1$, $\batchTmp=1$, $\neglocTmp=1$, and $\simcseTmp=1$. For \textit{\modelname-NCE-*}, the best combination is $\lrunsuper =0.0002$, $\neglocWeight=1$ and $\simcseWeight=1$.

On the fMoW dataset, the best hyperparameter combination for each model is similar to those on the iNat2018 dataset. The difference is $\lrunsuper$, and its  best value is $0.001$ for all models.

We further try to tune the location encoder hyperparameters for pretrained encoders, but found that the result parameters do not differ from those we got from previous hyperparameter tuning for \superonly{} model.
 \subsection{The Baseline Selection Criteria} \label{sec:baseline_sel_rule}

In the following, we will discuss the selection criteria we use to select baseline models, especially those \imageonly{} models shown in Figure \ref{fig:model-geo-ssl-detail}.

For the iNat2018 dataset, we only use \imageonly{} (ImageNet) and did not include some baselines such as \imageonly{} (Tile2Vec), \imageonly{} (Geo-SSL), and \imageonly{} (GeoKR) in Table \ref{tab:unsuper_eval_in18}, because they are not applicable:
\setlist{nolistsep}
\begin{itemize}[leftmargin=*]
    \setlength\itemsep{0.3em}
    \item \imageonly{} (Tile2Vec) assumes geospatially nearby remote sensing (RS) images are similar in the image embedding space. This assumption does not work for species images. Two bird from different species can locate nearby each other.
    \item \imageonly{} (Geo-SSL) needs to use RS images taken at the same location at different times as positive samples for self-supervised training. This idea does not work for species images either.
    \item \imageonly{} (GeoKR) requires geographically co-located land use maps which adds additional information, which makes it an unfair comparison.
\end{itemize}

For the fMoW dataset, we select \imageonly{} (Tile2Vec) and \imageonly{} (Geo-SSL) as two \imageonly{} baselines because:
\setlist{nolistsep}
\begin{itemize}[leftmargin=*]
    \setlength\itemsep{0.3em}
    \item  \imageonly{} (ImageNet) shows significantly lower performance than \imageonly{} (Geo-SSL) on fMoW according to \citet{ayush2020selfsup}. So we did not compare with it but used \imageonly{} (Geo-SSL) as the strong baseline. 
    \item \imageonly{} (GeoKR) requires additional global land use maps which leads to unfair comparison. 
    \item For \imageonly{} (Tile2Vec), its assumption is very weak in the fMoW dataset. In the original Tile2Vec paper \citep{jean2019tile2vec}, they took a large RS image and extracted nearby RS tiles from it. Some of the nearby RS tiles are usually very close or even share some regions. In the fMoW dataset, the RS images are samples from different locations and the nearby RS images are rather far away. We assume the performance of \imageonly{} (Tile2Vec) should be poor on fMoW. The experiment results in Table \ref{tab:unsuper_eval_fmow} confirm our assumption.
\end{itemize}

\subsection{Additional Related Work} \label{sec:add_related}
\paragraph{Unsupervised Representation Learning} Unsupervised text encoding models such as transformer~\cite{vaswani2017attention, devlin2018bert} has been effectively utilized in many Natural Language Processing (NLP) tasks. 
At its core, a trained model encodes words into vector space representations based on their positions and context in the text. 
Following the success in NLP, there has been significant recent progress in unsupervised image pretraining \cite{he2020moco, caron2020unsup,ayush2020selfsup}. Interestingly almost all of them are based on certain form of contrastive learning \cite{hadsell2006mapping}, which helps to construct unsupervised classification objectives from continuous inputs such as images.
\citet{he2020moco} proposes Momentum Contrast (MoCo) for unsupervised visual representation learning.  To increase the number of negative examples in contrastive training, they uses a queue of multiple mini-batches.  
Similar strategy has been adopted in NLP \cite{gao2021simcse}.
To improve the encoding consistency between mini batches,  they make the target image encoder parameterizes a moving average of the query image encoder. 
In this work we are focusing on the pretraining of location encoder with a frozen image encoder. Our approach is very memory efficient (easily scaling up to 8192 batch size) and therefore avoid the need of multi-batch training.

\paragraph{Contrastive Learning}

A contrastive training loss takes a pair of inputs $(\bx_i,\bx_j)$ and minimizes the embedding distance when they are similar according to certain signal (e.g., from the same class, or generated from the same original examples) but maximizes the distance otherwise.
Common effective ways to construct contrastive loss include 1) data augmentation techniques, which create noise/augmented versions of original examples as positive sample pairs \cite{gutmann2010nce,he2020moco,chen20simclr,zbontar2021barlow,gao2021simcse}; 2) construct in-batch negative pairs using a large batch size  \cite{chen20simclr,zhang2020contrastive,radford2021clip}; 3) hard negative mining for supervised learning tasks \cite{karpukhin2020dpr,gao2021simcse}.
These techniques has been successfully applied to a variety of image tasks \cite{ chen20simclr,he2020moco,zbontar2021barlow}, text tasks \cite{mnih2012lm,mikolov2013distributed,karpukhin2020dpr,gao2021simcse}, and multi-model tasks \cite{zhang2020contrastive,jia2021scaling,radford2021clip,zhai2021lit}. 
However, contrastive learning has never been used to learn image-location alignment in a pre-training set-up.
\modelname~ adapts contrastive strategies that work well on text and image and apply to geo-location data in order to construct positive and negative sample pairs.

\paragraph{Contrastive Learning on Multimodal Data} 
Recently, contrastive learning has been utilized on multimodal data (e.g. text-image pairs) by systems such as ConVIRT \cite{zhang2020contrastive}, CLIP \cite{radford2021clip}, and ALIGN \cite{jia2021scaling}. 
Given a set of text-image pairs, the text  and image data can be encoded separately by a text encoder and an image encoder. 
The resulting text and image representations are contrasted against each other such that the correct language-vision aligment is learned \cite{zhai2021lit}.
After this self-supervised pretraining, these models can be directly used for zero-shot transfer tasks such as image classificaion, image-text retrieval, and etc. 
While both CLIP \cite{radford2021clip} and ALIGN \cite{jia2021scaling} proposed to train the image encoder and text encoder jointly from scratch during contrastive pre-training,
LiT \cite{zhai2021lit} has shown that locking the image encoder that is initialized by a pre-trained model, while training text encoder from scratch during image-text contrastive pre-training can significantly improve the model performance on multiple downstream tasks.

\paragraph{Machine Learning on Spatial Data}
Recently, numerous studies have shown that appropriately incorporating (geo)spatial information into the learning framework can significantly improve the model performance on variety of geospatial tasks. 
Just to name a few, these tasks include species fine-grained recognition~\cite{chu2019geo,mac2019presence,mai2022sphere2vec},
ground-level image classification \cite{tang2015improving}, 
Point of Interest (POI) facade image classification \cite{yan2018xnet+},
POI type classification \cite{mai2020multiscale}, 
remote sensing (RS) image classification\footnote{Although remote sensing images can be largely regarded as geospatial data, here, we refer to the work which considers the geo-locations or timestamps of those RS images for ML model design instead of treating RS image classification as a pure computer vision task.} \cite{christie2018functional,ayush2020selfsup,manas2021seasonal}, 
poverty prediction \cite{jean2016combining,jean2019tile2vec}, 
land use classification \cite{jean2019tile2vec,ayush2020selfsup}, 
satellite image super-resolution \cite{he2021spatial}, 
and geographic question answering \cite{mai2020se,scheider2021geo}.
Despite all these success stories, these works either directly utilize spatial data in a supervised learning framework \cite{tang2015improving,christie2018functional,chu2019geo,mac2019presence,mai2020se,mai2020multiscale,mai2022sphere2vec}, or incorporate spatial data in an implicit manner in the unsupervised/self-supervised pre-training stage \cite{jean2019tile2vec,ayush2020selfsup,he2021spatial,manas2021seasonal,li2021geographical}.
The former cannot utilize massive unlabeled (geo)spatial datasets and performs poorly in a few-shot learning setting.
The latter only utilizes spatial data in the pre-training stage but ignores them at the model inference time so that the model performance at the inference time can be suboptimal.

\subsection{The Spatial Distribution of iNat2018 and fMoW dataset}  \label{sec:inat18_data_stat}

\begin{figure*}[ht!]
	\centering \vspace*{-0.2cm}
	\begin{subfigure}[b]{0.33\textwidth}  
		\centering 
		\includegraphics[width=\textwidth]{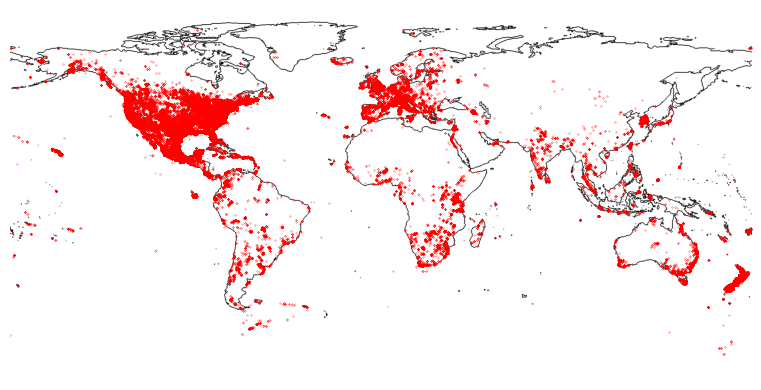}
		\vspace*{-0.9cm}
		\caption[]{{ Spatial Distribution of iNat2018 Train Data}}    
		\label{fig:inat18_train_locs}
	\end{subfigure}
	\begin{subfigure}[b]{0.33\textwidth}  
		\centering 
		\includegraphics[width=\textwidth]{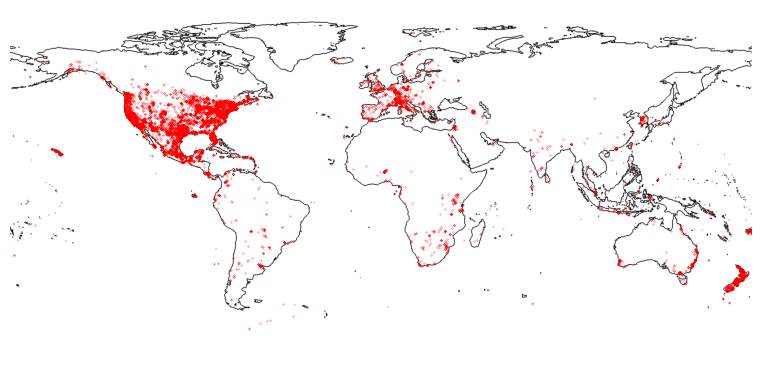}
		\vspace*{-0.9cm}
		\caption[]{{ Spatial Distribution of iNat2018 5\% Fine-Tune Data}}    
		\label{fig:inat18_train_sample_locs}
	\end{subfigure}
	\begin{subfigure}[b]{0.33\textwidth}  
		\centering 
		\includegraphics[width=\textwidth]{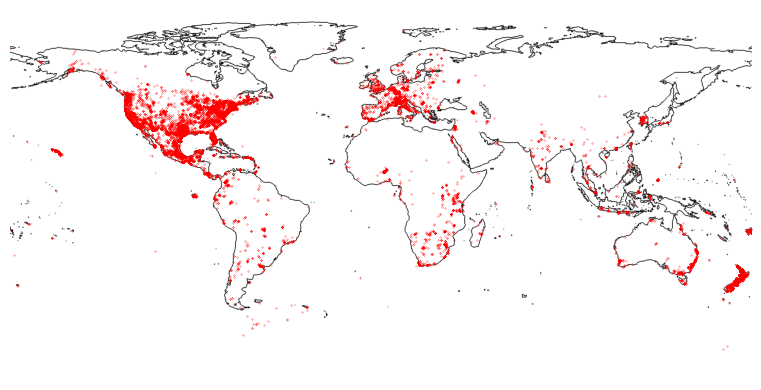}
		\vspace*{-0.9cm}
		\caption[]{{ Spatial Distribution of iNat2018 Validation Data}}    
		\label{fig:inat18_val_locs}
	\end{subfigure}
	\hfill
	\begin{subfigure}[b]{0.33\textwidth}  
		\centering 
		\includegraphics[width=\textwidth]{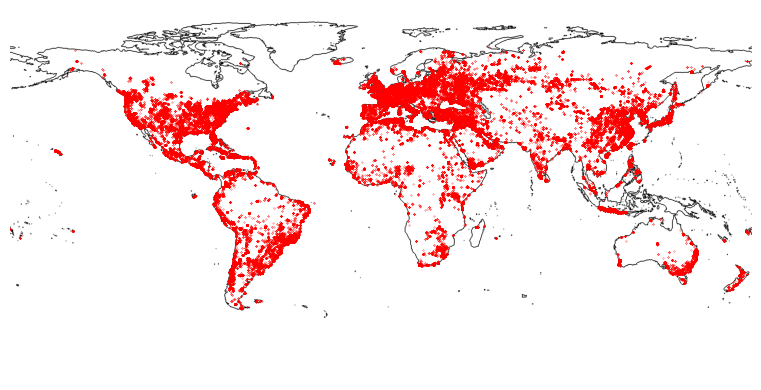}
		\vspace*{-0.9cm}
		\caption[]{{ Spatial Distribution of fMoW Train Data}}    
		\label{fig:fmow_train_locs}
	\end{subfigure}
	\begin{subfigure}[b]{0.33\textwidth}  
		\centering 
		\includegraphics[width=\textwidth]{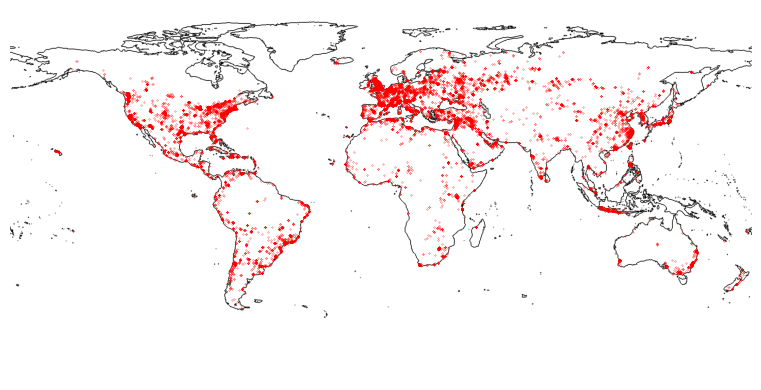}
		\vspace*{-0.9cm}
		\caption[]{{ Spatial Distribution of fMoW 5\% Fine-Tune Data}}    
		\label{fig:fmow_train_sample_locs}
	\end{subfigure}
	\begin{subfigure}[b]{0.33\textwidth}  
		\centering 
		\includegraphics[width=\textwidth]{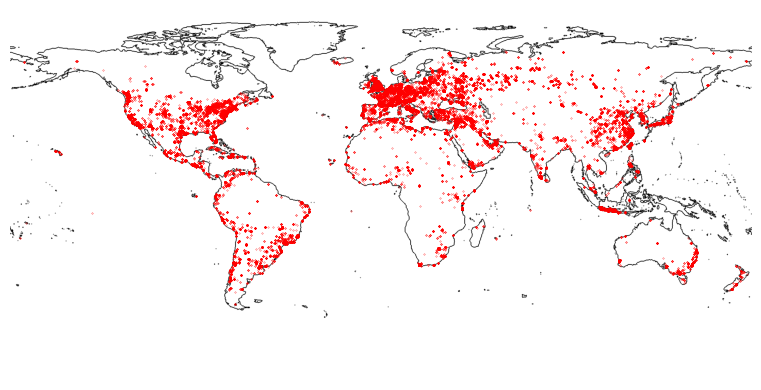}
		\vspace*{-0.9cm}
		\caption[]{{ Spatial Distribution of fMoW Validation Data}}    
		\label{fig:fmow_val_locs}
	\end{subfigure}
\caption{
	The spatial distribution of training, few-shot fine-tuning, and validation datasets for iNat2018 and fMoW.
	} 
	\label{fig:dataset_stat}
\end{figure*}

\end{document}